\documentclass[10pt,twocolumn,letterpaper]{article}

\usepackage{iccv}
\usepackage{times}
\usepackage{epsfig}
\usepackage{graphicx}
\usepackage{amsmath}
\usepackage{amssymb}
\usepackage{lib}
\usepackage{multirow}
\usepackage{float}
\usepackage{xcolor,colortbl}
\usepackage{makecell}
\usepackage[]{algorithm2e}

\usepackage[pagebackref=true,breaklinks=true,letterpaper=true,colorlinks,bookmarks=false]{hyperref}

\iccvfinalcopy % *** Uncomment this line for the final submission

\def\iccvPaperID{2473} % *** Enter the ICCV Paper ID here

\ificcvfinal\fi

\usepackage{enumitem}

\newlength\savewidth\newcommand\shline{\noalign{\global\savewidth\arrayrulewidth
  \global\arrayrulewidth 1pt}\hline\noalign{\global\arrayrulewidth\savewidth}}

\newcommand{\tablestyle}[2]{\setlength{\tabcolsep}{#1}\renewcommand{\arraystretch}{#2}\centering\footnotesize}

\newcommand{\ourpar}[1]{\vspace{1mm}\noindent\textbf{#1}}

\definecolor{Gray}{gray}{0.95}
\definecolor{Gray2}{rgb}{0.4, 0.4, 0.4}
\definecolor{LightCyan}{rgb}{0.88,1,1}
\definecolor{Lavender}{rgb}{0.909, 0.909, 0.95}

\newcommand{\Fone}[1]{F1$^{#1}$}

\newcolumntype{a}{>{\columncolor{Lavender}}c}
\newcolumntype{b}{>{\columncolor{white}}c}
\newcolumntype{T}{>{\footnotesize}c} % define a new column type for \tiny

\newcommand{\LL}{\mathcal{L}}
\DeclareMathOperator{\relu}{ReLU}

\newcommand{\APbox}{AP$^{\text{box}}$~}
\newcommand{\APmask}{AP$^{\text{mask}}$~}
\newcommand{\APmesh}{AP$^{\text{mesh}}$~}

\begin{document}

\title{Mesh R-CNN}

\author{
 Georgia Gkioxari \quad Jitendra Malik \quad Justin Johnson \vspace{3mm}\\
 Facebook AI Research (FAIR) \vspace{-2mm}
}
\maketitle
%\thispagestyle{empty}

%%%%%%%%% ABSTRACT
\begin{abstract}
  Rapid advances in 2D perception have led to systems that accurately detect objects in real-world images. However, these systems make predictions in 2D, ignoring the 3D structure of the world. Concurrently, advances in 3D shape prediction have mostly focused on synthetic benchmarks and isolated objects. We unify advances in these two areas. We propose a system that detects objects in real-world images and produces a triangle mesh giving the full 3D shape of each detected object. Our system, called Mesh R-CNN, augments Mask R-CNN with a mesh prediction branch that outputs meshes with varying topological structure by first predicting coarse voxel representations which are converted to meshes and refined with a graph convolution network operating over the mesh's vertices and edges. We validate our mesh prediction branch on ShapeNet, where we outperform prior work on single-image shape prediction. We then deploy our full Mesh R-CNN system on Pix3D, where we jointly detect objects and predict their 3D shapes. Project page: \small{\url{https://gkioxari.github.io/meshrcnn/}}.

\end{abstract}

%%%%%%%%% BODY TEXT
\vspace{-4mm}
\section{Introduction}
\seclabel{sec:Intro}

The last few years have seen rapid advances in 2D object recognition.
We can now build systems that accurately recognize objects~\cite{He2016,Krizhevsky2012,Simonyan2015,Szegedy2015},
localize them with 2D bounding boxes~\cite{Girshick2014,Ren2015} or masks~\cite{he2017maskrcnn},
and predict 2D keypoint positions~\cite{Cao2017,he2017maskrcnn,Toshev2014}
in cluttered, real-world images.
Despite their impressive performance, these systems ignore one critical fact:
that the world and the objects within it are 3D and extend beyond the $XY$ image plane.

At the same time, there have been significant advances in 3D shape understanding with deep networks.
A menagerie of network architectures have been developed for different 3D shape representations,
such as voxels~\cite{choy2016r2n2}, pointclouds~\cite{fan2017point}, and meshes~\cite{wang2018pixel2mesh};
each representation carries its own benefits and drawbacks. However, this diverse and creative set of
techniques has primarily been developed on synthetic benchmarks such as ShapeNet~\cite{shapenet} consisting
of rendered objects in isolation, which are dramatically less complex than natural-image benchmarks used
for 2D object recognition like ImageNet~\cite{Russakovsky2015} and COCO~\cite{Lin2014a}.

\begin{figure}
  \centering
  \begin{minipage}{0.49\linewidth} \centering Input Image \end{minipage}
    \begin{minipage}{0.49\linewidth} \centering 2D Recognition \end{minipage} \\*[1mm]
  \includegraphics[width=1.0\linewidth]{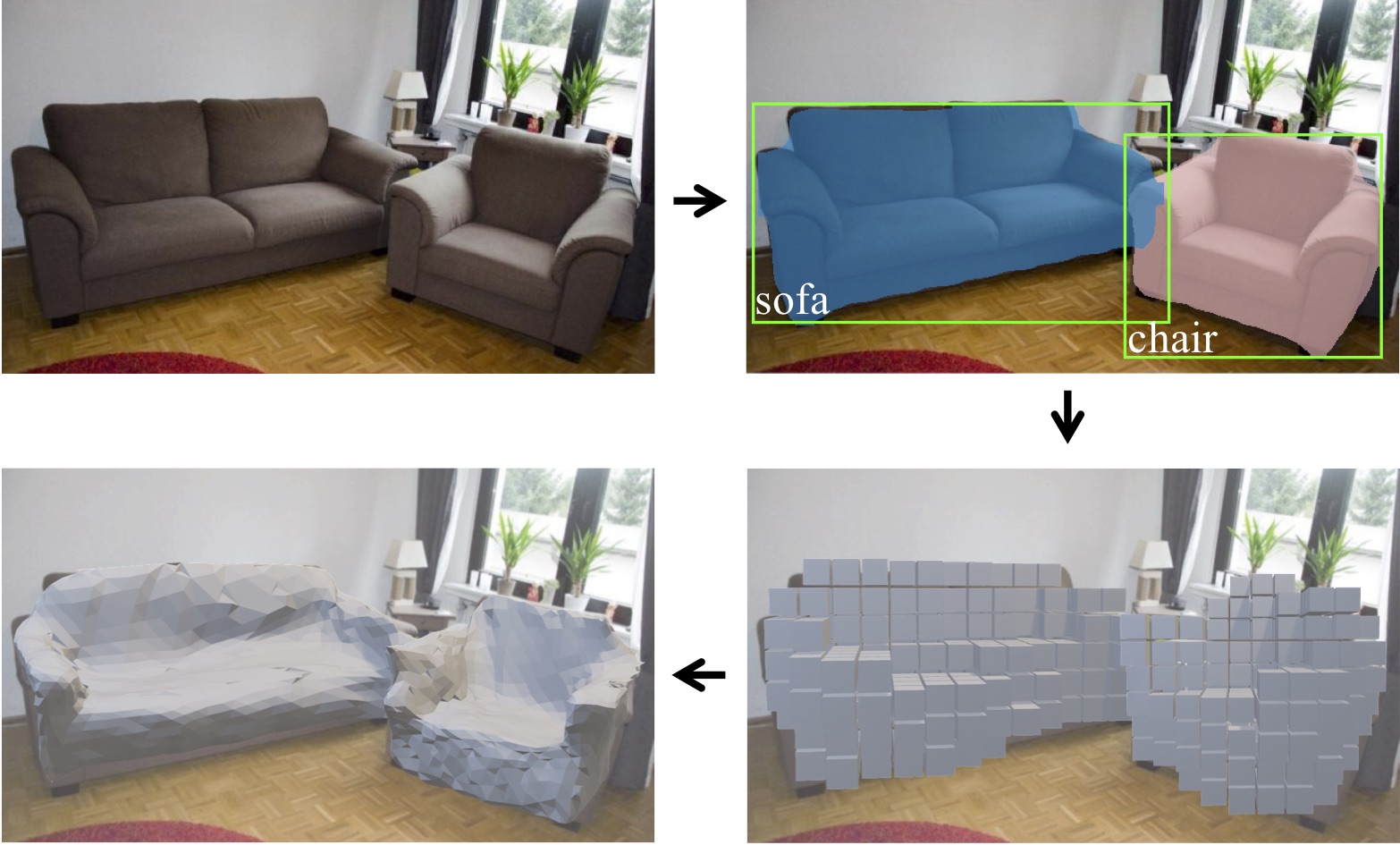} \\
  \begin{minipage}{0.49\linewidth} \centering 3D Meshes \end{minipage}
  \begin{minipage}{0.49\linewidth} \centering 3D Voxels \end{minipage}
  \vspace{2mm}
  \caption[Caption for teaser]{
    Mesh R-CNN takes an input image, predicts object instances in that image and infers their 3D shape.
    To capture diversity in geometries and topologies, it first predicts coarse voxels which are refined
    for accurate mesh predictions.
  }
   \label{fig:fig1}
   \vspace{-2mm}
\end{figure}

We believe that the time is ripe for these hitherto distinct research directions to be combined.
We should strive to build systems that (like current methods for 2D perception) can operate on unconstrained
real-world images with many objects, occlusion, and diverse lighting conditions but that (like current methods
for 3D shape prediction) do not ignore the rich 3D structure of the world.

In this paper we take an initial step toward this goal. We draw on state-of-the-art methods for 2D
perception and 3D shape prediction to build a system which inputs a real-world RGB image, detects
the objects in the image, and outputs a category label, segmentation mask, and a 3D triangle mesh
giving the full 3D shape of each detected object.

Our method, called \emph{Mesh R-CNN}, builds on the state-of-the-art Mask R-CNN~\cite{he2017maskrcnn}
system for 2D recognition, augmenting it with a \emph{mesh prediction branch} that outputs
high-resolution triangle meshes.

\begin{figure}[t!]
  \centering
  \includegraphics[width=1.0\linewidth]{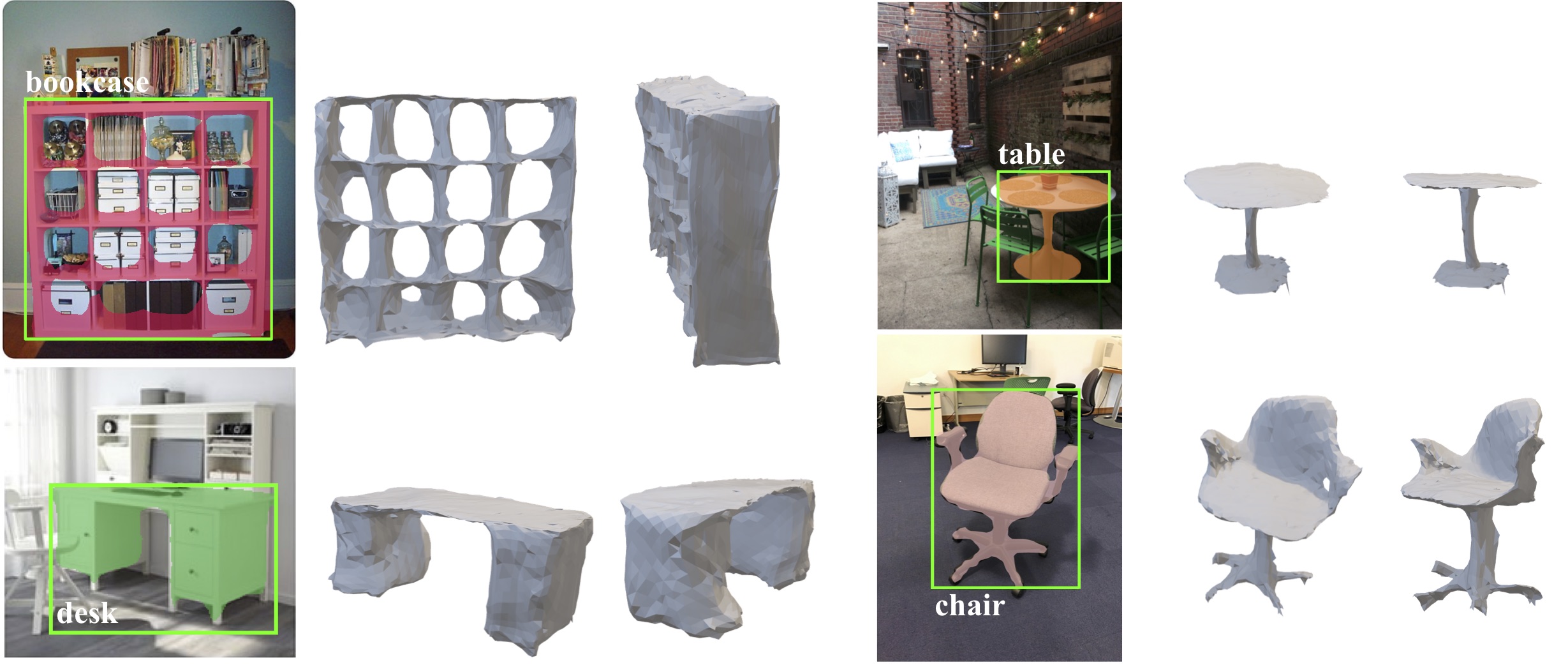}
  \caption{
    Example predictions from Mesh R-CNN on Pix3D. Using initial voxel predictions allows our outputs
    to vary in topology; converting these predictions to meshes and refining them allows us to 
    capture fine structures like tabletops and chair legs.
  }
  \vspace{-3mm}
  \label{fig:preds-teaser}
\end{figure}

Our predicted meshes must be able to capture the 3D structure of diverse, real-world objects.
Predicted meshes should therefore dynamically vary their complexity, topology, and geometry 
in response to varying visual stimuli.  However, prior work on mesh prediction with
deep networks~\cite{kanazawa2018learning,smith2019geometrics,wang2018pixel2mesh} has been constrained
to deform from fixed mesh templates, limiting them to fixed mesh topologies. As shown in
Figure~\ref{fig:fig1}, we overcome this limitation by utilizing multiple 3D shape representations:
we first predict coarse \emph{voxelized} object representations, which are converted to meshes and
refined to give highly accurate mesh predictions. As shown in Figure~\ref{fig:preds-teaser}, this
hybrid approach allows Mesh R-CNN to output meshes of arbitrary topology while also capturing fine
object structures.

We benchmark our approach on two datasets.
First, we evaluate our mesh prediction branch on ShapeNet~\cite{shapenet},
where our hybrid approach of voxel prediction and mesh refinement outperforms prior work by a large margin.
Second, we deploy our full Mesh R-CNN system on the recent Pix3D dataset~\cite{pix3d} which aligns
395 models of IKEA furniture to real-world images featuring diverse scenes, clutter, and occlusion.
To date Pix3D has primarily been used to evalute shape predictions for models
trained on ShapeNet, using perfectly cropped, unoccluded image segments~\cite{lmnet2018, pix3d, shapehd},
or synthetic rendered images of Pix3D models~\cite{genre}. In contrast, using Mesh R-CNN we are the first
to train a system on Pix3D which can jointly detect objects of all categories and estimate their full 3D shape.

\section{Related Work}
\seclabel{sec:Related}

\setlength{\parskip}{0pt}

Our system inputs a single RGB image and outputs a set of detected object instances, with a triangle mesh for each object. 
Our work is most directly
related to recent advances in 2D object recognition and 3D shape prediction.
We also draw more broadly from work on other 3D perception tasks.

\ourpar{2D Object Recognition}
Methods for 2D object recognition vary both in the type of information predicted per object, and in
the overall system architecture. Object detectors output per-object bounding boxes and category
labels~\cite{Girshick2015,Girshick2014,lin2017focal,Liu2016,Redmon2016,Ren2015};
Mask R-CNN~\cite{he2017maskrcnn} additionally outputs instance segmentation masks.
Our method extends this line of work to output a full 3D mesh per object.

\ourpar{Single-View Shape Prediction} 
Recent approaches use a variety of shape representations for single-image 3D reconstruction.
Some methods predict the orientation~\cite{Fouhey13,Hoiem2005GeometricCF} or
3D pose~\cite{3DRCNN_CVPR18,pavlakos17object3d,vpsKpsTulsianiM15} of known shapes.
Other approaches predict novel 3D shapes as
sets of 3D points~\cite{fan2017point,lin2018learning},
patches~\cite{groueix2018papier,wang2018adaptive}, or geometric
primitives~\cite{Fidler2012,tian2019learning,tulsiani2017learning};
others use deep networks to model signed distance functions~\cite{mescheder2018occupancy}.
These methods can flexibly represent complex shapes, 
but rely on post-processing to extract watertight mesh outputs.

Some methods predict regular voxel grids~\cite{choy2016r2n2,wu2017marrnet,wu2016learning}; while
intuitive, scaling to high-resolution outputs requires complex
octree~\cite{riegler2017octnet,tatarchenko2017octree} or nested
shape architectures~\cite{richter2018matryoshka}.

Others directly output triangle meshes, but are constrained to deform from
fixed~\cite{smith2018multi,smith2019geometrics,wang2018pixel2mesh} or retrieved
mesh templates~\cite{rock2015completing}, limiting the topologies they can represent.

Our approach uses a hybrid of voxel prediction and mesh deformation, enabling high-resolution
output shapes that can flexibly represent arbitrary topologies.

Some methods reconstruct 3D shapes without 3D
annotations~\cite{kanazawa2018learning,kato2018neural,rezende2016unsupervised,drcTulsiani17,yan2016perspective}.
This is an important direction, but at present we consider only the
fully supervised case due to the success of strong supervision for 2D perception.

\ourpar{Multi-View Shape Prediction} There is a broad line of work on multi-view reconstruction of objects and scenes, from classical binocular stereo~\cite{hartley2003multiple, scharstein2002taxonomy} to using shape priors~\cite{Bao2013,blanz1999,dame2013,hane2014} and modern learning techniques~\cite{kar2017learning,kendall2017end,schmidt2017self}. In this work, we focus on single-image shape reconstruction.

\ourpar{3D Inputs}
Our method inputs 2D images and predicts semantic labels and 3D shapes.
Due to the increasing availabilty of depth sensors, there has been growing interest in methods
predicting semantic labels from 3D inputs such as
RGB-D images~\cite{gupta2014learning, DeepSlidingShapes} and
pointclouds~\cite{graham2018sparse,li2018pointcnn,qi2017pointnetplusplus,su18splatnet,maxim2018tang}.
We anticipate that incorporating 3D inputs into our method could improve the fidelity of our
shape predictions.

\ourpar{Datasets}
Advances in 2D perception have been driven by large-scale annotated datasets such as
ImageNet~\cite{Russakovsky2015} and COCO~\cite{Lin2014a}. Datasets for 3D shape prediction have
lagged their 2D counterparts due to the difficulty of collecting 3D annotations.
ShapeNet~\cite{shapenet} is a large-scale dataset of CAD models which are rendered to give synthetic images.
The IKEA dataset~\cite{lim2013ikea} aligns CAD models of IKEA objects to real-world images;
Pix3D~\cite{pix3d} extends this idea to a larger set of images and models.
Pascal3D~\cite{xiang_wacv14} aligns CAD models to real-world images, but it is unsuitable for shape
reconstruction since its train and test sets share the same small set of models.
KITTI~\cite{Geiger2013IJRR} annotates outdoor street scenes with 3D bounding boxes, but does not
provide shape annotations.

\begin{figure*}[ht!]

  \centering
  \includegraphics[width=0.97\linewidth]{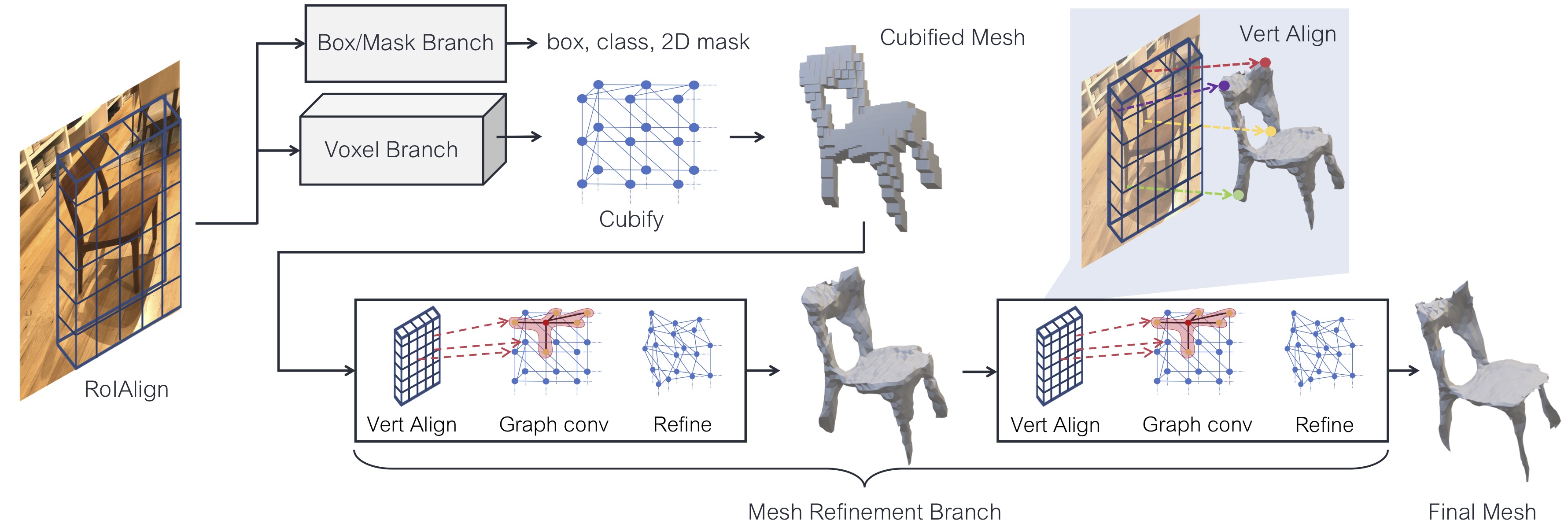}
  \caption{
    System overview of Mesh R-CNN. We augment Mask R-CNN with 3D shape inference. The \emph{voxel branch} predicts a coarse shape for each detected object which is further deformed with a sequence of refinement stages in the \emph{mesh refinement branch}.
  }
  \label{fig:system}
  \vspace{-2mm}
\end{figure*}

\section{Method}
\label{sec:Approach}
Our goal is to design a system that inputs a single image, detects all objects,
and outputs a category label, bounding box, segmentation mask and 3D triangle mesh for each detected
object. Our system must be able to handle cluttered real-world images, and must be trainable
end-to-end. Our output meshes should not be constrained to a fixed topology in order to
accommodate a wide variety of complex real-world objects. 
We accomplish these goals by marrying state-of-the-art 2D perception with 3D shape prediction.

Specifically, we build on Mask R-CNN~\cite{he2017maskrcnn}, a state-of-the-art 2D perception system.
Mask R-CNN is an end-to-end region-based object detector.
It inputs a single RGB image and outputs a bounding box, category label, and segmentation mask
for each detected object.
The image is first passed through a \emph{backbone network}
(\eg ResNet-50-FPN~\cite{Lin2017fpn}); 
next a \emph{region proposal network} (RPN) ~\cite{Ren2015} gives object proposals
which are processed with object classification and mask prediction branches.

Part of Mask R-CNN's success is due to \texttt{RoIAlign} which extracts region
features from image features while maintaining alignment between the input image and
features used in the final prediction branches.
We aim to maintain similar feature alignment when predicting 3D shapes.

We infer 3D shapes with a novel mesh predictor, comprising a \emph{voxel branch} and a \emph{mesh refinement branch}.
The voxel branch first estimates a coarse 3D voxelization of an object, which is converted to an initial triangle mesh.
The mesh refinement branch then adjusts the vertex positions of this initial mesh using a sequence of graph convolution layers operating over the edges of the mesh.

The voxel branch and mesh refinement branch are homologous to the existing box and mask branches of Mask R-CNN. All take as input image-aligned features corresponding to RPN proposals. The voxel and mesh losses, described in detail below, are added to the box and mask losses and the whole system is trained end-to-end. The output is a set of boxes along with their predicted object scores, masks and 3D shapes. We call our system \emph{Mesh R-CNN}, which is illustrated in Figure~\ref{fig:system}.

We now describe our mesh predictor, consisting of the voxel branch and mesh refinement branch, along with its associated losses in detail.
 
\subsection{Mesh Predictor}
At the core of our system is a mesh predictor which receives convolutional features
aligned to an object's bounding box and predicts a triangle mesh giving the object's full 3D
shape.
Like Mask R-CNN, we maintain correspondence between the input image and features used at all
stages of processing via region- and vertex-specific alignment operators (\texttt{RoIAlign} and
\texttt{VertAlign}).
Our goal is to capture instance-specific 3D shapes of all objects in an image. Thus, each predicted mesh must have instance-specific \emph{topology} (genus, number of vertices,
faces, connected components) and \emph{geometry} (vertex positions).

We predict varying mesh topologies by deploying a sequence of shape inference operations.
First, the \emph{voxel branch} makes bottom-up voxelized predictions of each object's shape,
similar to Mask R-CNN's mask branch. These predictions are converted into meshes and adjusted by the
\emph{mesh refinement head}, giving our final predicted meshes. 

The output of the mesh predictor is a \emph{triangle mesh} $T\!=\!(V, F)$ for each object.
$V\!=\!\{v_i\!\in\!\mathbb{R}^3\}$ is the set of vertex positions and
$F\!\subseteq\!V\hspace{-1mm}\times\hspace{-1mm}V\hspace{-1mm}\times\hspace{-1mm}V$ is a set of triangular faces.

\subsubsection{Voxel Branch}
The \emph{voxel branch} predicts a grid of voxel occupancy probabilities giving the course 3D
shape of each detected object.
It can be seen as a 3D analogue of Mask R-CNN's mask prediction branch: rather than
predicting a $M\times M$ grid giving the object's shape in the image plane, we instead predict
a $G\times G\times G$ grid giving the object's full 3D shape.

Like Mask R-CNN, we maintain correspondence between input features and predicted
voxels by applying a small fully-convolutional network~\cite{Long2015} to the input feature map
resulting from \texttt{RoIAlign}. This network produces a feature map with $G$
channels giving a column of voxel occupancy scores for each position in the input.

\begin{figure}[t!]
  \centering
  \hspace{0.1\linewidth}
  \begin{minipage}{0.5\linewidth}\centering \textbf{World Space} \end{minipage}
  \begin{minipage}{0.35\linewidth}\centering \textbf{Prediction Space} \end{minipage}
  \includegraphics[width=0.9\linewidth]{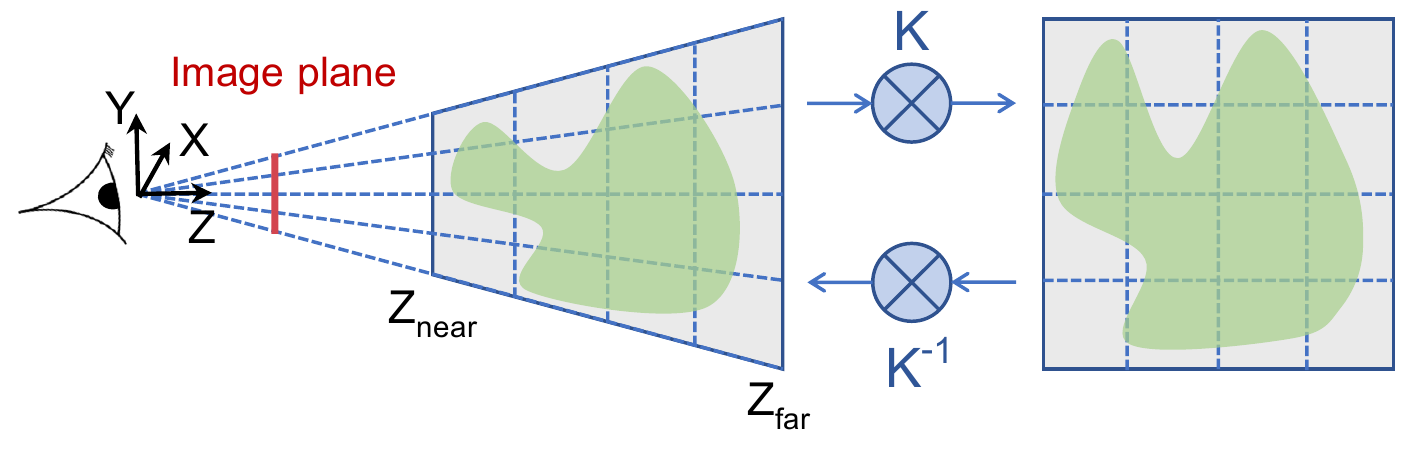}
  \caption{
    Predicting voxel occupancies aligned to the image plane requires an irregularly-shaped voxel grid.
    We achieve this effect by making voxel predictions in a space that is transformed by the
    camera's (known) intrinsic matrix $K$. Applying $K^{-1}$ transforms our predictions back
    to world space. This results in frustum-shaped voxels in world space.
  }
  \label{fig:voxel_grid}
  \vspace{-2mm}
\end{figure}

Maintaining pixelwise correspondence between the image and our predictions is complex in 3D
since objects become smaller as they recede from the camera. As shown in
Figure~\ref{fig:voxel_grid}, we account for this by using the camera's (known) intrinsic matrix to
predict frustum-shaped voxels.

\ourpar{Cubify: Voxel to Mesh}
The voxel branch produces a 3D grid of occupancy probabilities giving the coarse shape of an
object. In order to predict more fine-grained 3D shapes, we wish to convert these voxel predictions
into a triangle mesh which can be passed to the mesh refinement branch.

We bridge this gap with an operation called \texttt{cubify}. It inputs voxel occupancy
probabilities and a threshold for binarizing voxel occupancy. 
Each occupied voxel is replaced with a cuboid triangle mesh with
8 vertices, 18 edges, and 12 faces. Shared vertices and edges between adjacent occupied voxels
are merged, and shared interior faces are eliminated. This results in a watertight mesh
whose topology depends on the voxel predictions.

\texttt{Cubify} must be efficient and batched. This is not trivial and we provide technical implementation details of how we achieve this in Appendix~\ref{app:cubify}.
Alternatively marching cubes~\cite{lorensen1987marching} could extract an isosurface from the voxel grid, but is significantly more complex.

\ourpar{Voxel Loss} The voxel branch is trained to minimize the binary cross-entropy between predicted
voxel occupancy probabilities and true voxel occupancies.

\subsubsection{Mesh Refinement Branch}
The cubified mesh from the voxel branch only provides a coarse 3D shape, and it cannot accurately
model fine structures like chair legs.
The \emph{mesh refinement branch} processes this initial cubified mesh,
refining its vertex positions with a sequence of refinement stages. Similar to \cite{wang2018pixel2mesh}, each refinement stage
consists of three operations:
\emph{vertex alignment}, which extracts image features for vertices;
\emph{graph convolution}, which propagates information along mesh edges; and
\emph{vertex refinement}, which updates vertex positions.
Each layer of the network maintains a 3D position $v_i$ and a feature vector $f_i$ for each mesh vertex.

\ourpar{Vertex Alignment} yields an image-aligned feature vector for each mesh vertex\footnote{
  Vertex alignment is called \emph{perceptual feature pooling} in \cite{wang2018pixel2mesh}}.
We use the camera's intrinsic matrix to project each vertex onto
the image plane. Given a feature map, 
we compute a bilinearly interpolated
image feature at each projected vertex position~\cite{Jaderberg2015}. 

In the first stage of the mesh refinement branch, \texttt{VertAlign} outputs an initial
feature vector for each vertex. In subsequent stages, the \texttt{VertAlign} output is
concatenated with the vertex feature from the previous stage.

\ourpar{Graph Convolution }\cite{kipf2017semi} propagates information along mesh
edges. Given input vertex features $\{f_i\}$ it computes updated features
{$\small f_i' = \relu\big(W_0 f_i + \sum_{j\in\mathcal{N}(i)} W_1f_j\big)$}
where $\mathcal{N}(i)$ gives the $i$-th vertex's neighbors in the mesh,
and $W_0$ and
$W_1$ are learned weight matrices. Each stage of the mesh refinement branch uses several graph
convolution layers to aggregate information over local mesh regions.

\ourpar{Vertex Refinement} computes updated vertex positions
$v_i' = v_i + \tanh(W_{vert}\left[f_i; v_i\right])$
where $W_{vert}$ is a learned weight matrix. This updates the mesh
\emph{geometry}, keeping its \emph{topology} fixed.
Each stage of the mesh refinement branch terminates with vertex refinement,
producing an intermediate mesh output which is further refined by the next stage.

\ourpar{Mesh Losses}
Defining losses that operate natively on triangle meshes is challenging,
so we instead use loss functions defined over a finite set of points. 
We represent a mesh with a pointcloud by densely sampling its surface. 
 Consequently, a pointcloud loss approximates a loss over shapes.

Similar to \cite{smith2019geometrics}, we use a differentiable \emph{mesh sampling} operation
to sample points (and their normal vectors) uniformly from the surface of a mesh. 
To this end, we implement an efficient batched sampler; see Appendix~\ref{app:meshsampling} for details. 
We use this operation to sample a pointcloud $P^{gt}$
from the ground-truth mesh, and a pointcloud $P^i$ from each intermediate mesh prediction
from our model.

Given two pointclouds $P$, $Q$ with normal vectors,
let $\Lambda_{P,Q} = \{(p, \arg\min_q\|p-q\|) : p\in P\}$
be the set of pairs $(p, q)$ such that $q$ is the nearest neighbor of $p$ in $Q$, and let
$u_p$ be the unit normal to point $p$. The \emph{chamfer distance} between
pointclouds $P$ and $Q$ is given by
\begin{equation}
  \label{eq:chamfer}
  \footnotesize
  \LL_{\textrm{cham}}(P, Q) =
    |P|^{-1} \hspace{-4mm} \sum_{(p,q)\in\Lambda_{P,Q}}\hspace{-3mm}\|p-q\|^2
  + |Q|^{-1} \hspace{-4mm} \sum_{(q,p)\in\Lambda_{Q,P}}\hspace{-3mm}\|q-p\|^2
\end{equation}
and the (absolute) \emph{normal distance} is given by
\begin{equation}
  \label{eq:normal}
  \footnotesize
  \LL_{\textrm{norm}}(P, Q) =
    -|P|^{-1} \hspace{-4mm} \sum_{(p,q)\in\Lambda_{P,Q}}\hspace{-3mm}|u_p \cdot u_q|
    -|Q|^{-1} \hspace{-4mm} \sum_{(q,p)\in\Lambda_{Q,P}}\hspace{-3mm}|u_q\cdot u_p|.
\end{equation}
The chamfer and normal distances penalize mismatched positions and normals between
two pointclouds, but minimizing these distances alone results in degenerate meshes
(see Figure~\ref{fig:best-vs-pretty}).
High-quality mesh predictions require additional \emph{shape regularizers}:
To this end we use an \emph{edge loss}
$\LL_{\textrm{edge}}(V, E)=\frac{1}{|E|}\sum_{(v,v')\in E}\|v-v'\|^2$ where $E\subseteq V\times V$
are the edges of the predicted mesh. 
Alternatively, a Laplacian loss~\cite{laplacian} also imposes smoothness constraints.

The \emph{mesh loss} of the $i$-th stage is a weighted sum of
$\LL_{\textrm{cham}}(P^i, P^{gt})$, $\LL_{\textrm{norm}}(P^i, P^{gt})$ and $\LL_{\textrm{edge}}(V^i, E^i)$. 
The mesh refinement branch is trained to minimize the mean of these losses across all refinement stages.

\section{Experiments}
\label{sec:Experiments}
% Snooping much?!

We benchmark our mesh predictor on ShapeNet~\cite{shapenet}, where we compare with
state-of-the-art approaches.
We then evaluate our full Mesh R-CNN for the task of 3D shape prediction \emph{in the wild} on the
challenging Pix3D dataset~\cite{pix3d}.

\subsection{ShapeNet}
\seclabel{shapenet}

ShapeNet~\cite{shapenet} provides a collection of 3D shapes, represented as
textured CAD models organized into semantic categories following
WordNet~\cite{wordnet}, and has been widely used as a benchmark for
3D shape prediction. We use the subset of ShapeNetCore.v1 and rendered images
from~\cite{choy2016r2n2}. Each mesh is rendered from up to 24 random viewpoints, giving
RGB images of size\ $137\times137$. We use the train / test splits provided by~\cite{wang2018pixel2mesh},
which allocate 35,011 models (840,189 images) to train and 8,757 models
(210,051 images) to test; models used in train and test are disjoint.
We reserve 5\% of the training models as a validation set.

The task on this dataset is to input a single RGB image of a rendered ShapeNet
model on a blank background, and output a 3D mesh for the object in the camera coordinate system. 
During training the system is supervised with pairs of images and meshes.

% split:  train
%   models:  33268
%   images:  798357
% split:  val
%   models:  1743
%   images:  41832
% split:  test
%   models:  8757
%   images:  210051
% 
% images per model:
%   min:  4
%   max:  24
%   mean:  23.995613233412538
%   median:  24.0

\ourpar{Evaluation}
We adopt evaluation metrics used in recent work~\cite{smith2018multi,smith2019geometrics,wang2018pixel2mesh}.
We sample 10k points uniformly at random from the surface of predicted and ground-truth meshes,
and use them to compute Chamfer distance (Equation~\ref{eq:chamfer}), Normal consistency,
(one minus Equation~\ref{eq:normal}), and \Fone{\tau} at various distance thresholds
$\tau$, which is the harmonic mean of the precision at $\tau$ (fraction of
predicted points within $\tau$ of a ground-truth point) and the recall at $\tau$ (fraction of
ground-truth points within $\tau$ of a predicted point).
Lower is better for Chamfer distance; higher is better for all other metrics.

With the exception of normal consistency, these metrics depend on the absolute
scale of the meshes.
In Table~\ref{table:shapenet-main} we follow \cite{wang2018pixel2mesh} and rescale by a factor of 0.57;
for all other results we follow \cite{fan2017point} and rescale so the longest edge of the ground-truth
mesh's bounding box has length 10.

\begin{table}
  \centering
  \tablestyle{3.5pt}{1.1}
  \begin{tabular}{l|ccc} 
    & Chamfer ($\downarrow$)
    & \Fone{\tau} ($\uparrow$)
    & \Fone{2\tau} ($\uparrow$) \\
    \shline
    N3MR~\cite{kato2018neural} & 2.629 & 33.80 & 47.72 \\
    3D-R2N2~\cite{choy2016r2n2} & 1.445 & 39.01 & 54.62 \\
    PSG~\cite{fan2017point} & 0.593 & 48.58 & 69.78 \\
    Pixel2Mesh~\cite{wang2018pixel2mesh}$^\dagger$ & 0.591 & 59.72 & 74.19 \\
    MVD~\cite{smith2018multi}& - & 66.39 & - \\
    GEOMetrics~\cite{smith2019geometrics} & - & 67.37 & - \\
    Pixel2Mesh~\cite{wang2018pixel2mesh}$^\ddagger$ & 0.463 & 67.89 & 79.88 \\
    \hline 
    % /mnt/vol/gfsai-east/ai-group/users/jcjohns/iccv19/sweeps/3-5-vox-mesh/task6/
    % Ours (Pretty) & 0.378 & 70.00 & 82.06 \\
    Ours (Best) & \bf{0.306} & \bf{74.84} & \bf{85.75}  \\ % 3-13-vox-mesh-48/task0
    Ours (Pretty) & 0.391 & 69.83 & 81.76 % 3-13-vox-mesh-48/task1
    % /mnt/vol/gfsai-east/ai-group/users/jcjohns/iccv19/sweeps/3-4-vox-mesh/task0/
    % Ours (Best) & \textbf{0.290} & \textbf{74.8}2 & \textbf{86.17} \\
  \end{tabular}
  \vspace{1mm}
  \caption{
    Single-image shape reconstruction results on ShapeNet,
    using the evaluation protocol from~\cite{wang2018pixel2mesh}.
    For \cite{wang2018pixel2mesh}, $^\dagger$ are results reported in their paper
    and $^\ddagger$ is the model released by the authors.
%    (mesh scaling factor of 0.57).
%    $^\ddagger$ indicates the released model by the authors
%    of \cite{wang2018pixel2mesh}. $\tau=10^{-4}$.
  }
  \vspace{-6mm}
  \label{table:shapenet-main}
\end{table}

%% ---- Justin's caption ------ %
%Single-image shape reconstruction results on ShapeNet,
%using the evaluation protocol from~\cite{wang2018pixel2mesh}.
%Rows marked with $^\star$ were reported by~\cite{wang2018pixel2mesh}
%and evaluate with a different sampling strategy from us;
%$^\dagger$ were reported in the respective publications and use
%different train / test splits;
%$^\ddagger$ indicates the pretrained model released by the authors
%of \cite{wang2018pixel2mesh} using our evaluation code, and is the
%best comparison with ours.

\ourpar{Implementation Details}
Our backbone feature extractor is ResNet-50 pretrained on ImageNet.
Since images depict a single object, the voxel branch receives the entire
\texttt{conv5\_3} feature map, bilinearly resized to $24\times24$, and
predicts a $48\times48\times48$ voxel grid. The \texttt{VertAlign} operator
concatenates features from \texttt{conv2\_3}, \texttt{conv3\_4}, \texttt{conv4\_6}, and \texttt{conv5\_3}
before projecting to a vector of dimension 128.
The mesh refinement branch has three stages, each with six graph convolution layers (of dimension 128) organized into three residual blocks.
We train for 25 epochs using
Adam~\cite{kingma2015adam} with learning rate $10^{-4}$ and 32 images per batch
on 8 Tesla V100 GPUs.
We set the \texttt{cubify} threshold to 0.2 and weight the losses with
$\lambda_{\textrm{voxel}}=1$, $\lambda_{\textrm{cham}}=1$, $\lambda_{\textrm{norm}}=0$, and $\lambda_{\textrm{edge}}=0.2$.

\ourpar{Baselines} We compare with previously published methods
for single-image shape prediction.
N3MR~\cite{kato2018neural} is a weakly supervised approach that fits a mesh via a differentiable renderer without 3D supervision. 3D-R2N2~\cite{choy2016r2n2} and MVD~\cite{smith2018multi} output voxel predictions. PSG~\cite{fan2017point} predicts point-clouds.
Appendix~\ref{app:occnet} additionally compares with OccNet~\cite{mescheder2018occupancy}.

Pixel2Mesh~\cite{wang2018pixel2mesh} predicts meshes by deforming and subdividing an initial ellipsoid. GEOMetrics~\cite{smith2019geometrics} extends \cite{wang2018pixel2mesh} with adaptive face subdivision.
Both are trained to minimize Chamfer distances; however \cite{wang2018pixel2mesh} computes it using predicted mesh vertices,
while \cite{smith2019geometrics} uses points sampled uniformly from predicted meshes. We adopt the latter as it
better matches test-time evaluation.
Unlike ours, these methods can only predict connected meshes of genus zero.

\newcommand{\meanstd}[2]{#1\scalebox{0.65}{$\pm#2$}}
\begin{table*}[ht]
  \centering
  \tablestyle{3pt}{1.1}
  \scalebox{0.95}{
  \begin{tabular}{cl|ccccccc|ccccccc}
    & & \multicolumn{7}{c|}{Full Test Set}
      & \multicolumn{7}{c}{Holes Test Set} \\
    & & Chamfer($\downarrow$) & Normal & \Fone{0.1} & \Fone{0.3} & \Fone{0.5} & $|V|$ & $|F|$
    & Chamfer($\downarrow$) & Normal & \Fone{0.1} & \Fone{0.3} & \Fone{0.5} & $|V|$ & $|F|$ \\
    \shline
    & Pixel2Mesh~\cite{wang2018pixel2mesh}$^\ddagger$
      & 0.205 & 0.736 & 33.7 & 80.9 & 91.7
        & \meanstd{2466}{0} & \meanstd{4928}{0}
      & 0.272 & 0.689 & 31.5 & 75.9 & 87.9
        & \meanstd{2466}{0} & \meanstd{4928}{0} \\
    \hline
     & Voxel-Only %($48^3$) % 3-3-voxel/task3
      & 0.916 & 0.595 & 7.7 & 33.1 & 54.9
        & \meanstd{1987}{936} & \meanstd{3975}{1876}
      & 0.760 & 0.592 & 8.2 & 35.7 & 59.5
        & \meanstd{2433}{925} & \meanstd{4877}{1856} \\
    \hline\hline
     \multirow{4}{*}{\rotatebox[origin=c]{90}{Best}}
     & Sphere-Init % 3-13-sphere-iso4/task0
      & \bf{0.132} & 0.711 & 38.3 & 86.5 & \bf{95.1}
        & \meanstd{2562}{0} & \meanstd{5120}{0}
      & 0.138 & 0.705 & 40.0 & 85.4 & 94.3
        & \meanstd{2562}{0} & \meanstd{5120}{0} \\
     & Pixel2Mesh$^+$ % 3-8-pix2mesh-head/task0
      & {\bf 0.132} & 0.707 & 38.3 & 86.6 & {\bf 95.1}
        & \meanstd{2562}{0} & \meanstd{5120}{0}
      & 0.137 & 0.696 & 39.3 & 85.5 & 94.4
        & \meanstd{2562}{0} & \meanstd{5120}{0} \\
    & Ours (light) % 4-09-three-stage/task2
      & 0.133 & 0.725 & {\bf 39.2} & {\bf 86.8} & {\bf 95.1}
        & \meanstd{1894}{925} & \meanstd{3791}{1855}
      & {\bf 0.130} & 0.723 & 41.6 & {\bf 86.7} & 94.8
        & \meanstd{2273}{899} & \meanstd{4560}{1805} \\
    & Ours %-48 % 3-13-vox-mesh-48/task0
      & 0.133 & {\bf 0.729} & 38.8 & 86.6 & {\bf 95.1}
        & \meanstd{1899}{928} & \meanstd{3800}{1861}
      & {\bf 0.130} & {\bf 0.725} & {\bf 41.7} & {\bf 86.7} & {\bf 94.9}
        & \meanstd{2291}{903} & \meanstd{4595}{1814} \\
    \hline\hline
    \multirow{4}{*}{\rotatebox[origin=c]{90}{Pretty}}
     & Sphere-Init % 3-13-sphere-iso4/task2
       & 0.175 & 0.718 & 34.5 & 82.2 & 92.9
         & \meanstd{2562}{0} & \meanstd{5120}{0}
       & 0.186 & 0.684 & 34.4 & 80.2 & 91.7
         & \meanstd{2562}{0} & \meanstd{5120}{0} \\
     & Pixel2Mesh$^+$ % 3-12-sphere/task0
       & 0.175 & \bf{0.727} & 34.9 & 82.3 & 92.9
         & \meanstd{2562}{0} & \meanstd{5120}{0}
      & 0.196 & 0.685 & 34.4 & 79.9 & 91.4
         & \meanstd{2562}{0} & \meanstd{5120}{0} \\
    & Ours (light) % 4-09-three-stage/task3
      & 0.176 & 0.699 & 34.8 & 82.4 & 93.1
        & \meanstd{1891}{924} & \meanstd{3785}{1853}
      & 0.178 & 0.688 & 36.3 & 82.0 & 92.4
        & \meanstd{2281}{895} & \meanstd{4576}{1798} \\
    & Ours % -48 % 3-13-vox-mesh-48/task1
     & {\bf 0.171} & 0.713 & {\bf 35.1} & {\bf 82.6} & {\bf 93.2}
       & \meanstd{1896}{928} & \meanstd{3795}{1861}
     & {\bf 0.171} & \bf{0.700} & {\bf 37.1} & {\bf 82.4} & {\bf 92.7}
       & \meanstd{2292}{902} & \meanstd{4598}{1812}
  \end{tabular}}
  \vspace{1mm}
  \caption{
    We report results both on the full ShapeNet test set (left), as well as a subset of the test set
    consisting of meshes with visible holes (right). We compare our full model with several ablated version: Voxel-Only omits the mesh refinement
    head, while Sphere-Init and Pixel2Mesh$^+$ omit the voxel head. We show results both for
    \emph{Best} models which optimize for metrics, as well as \emph{Pretty} models that strike a
    balance between shape metrics and mesh quality (see Figure~\ref{fig:best-vs-pretty});
    these two categories of models should not be compared. We also report the number of vertices $|V|$ and faces $|F|$ in predicted meshes (mean$_{\pm \textrm{std}}$). $^\ddagger$ refers to the released model by the authors. (per-instance average)
  }
  \label{table:shapenet-ablation}
\end{table*}

The training recipe and backbone architecture vary among prior work.
Therefore for a fair comparison with our method we also compare against
several ablated versions of our model (see Appendix~\ref{app:arch} for exact
details):
\vspace{-2mm}
\begin{itemize}[leftmargin=*]
  \setlength{\itemsep}{-1mm}
  \item Voxel-Only: A version of our method that terminates with the cubified meshes
    from the voxel branch.
  \item Pixel2Mesh$^+$: We reimplement Pixel2Mesh~\cite{wang2018pixel2mesh};
  we outperform their original model due to a deeper backbone, better training recipe,
  and minimizing Chamfer on sampled rather than vertex positions.
 \item Sphere-Init: Similar to Pixel2Mesh$^+$, but initializes from a high-resolution
  sphere mesh, performing three stages of vertex refinement without subdivision.
 \item Ours (light): Uses a smaller nonresidual mesh refinement branch with three
   graph convolution layers per stage. We will adopt this lightweight design on Pix3D.
\end{itemize}
\vspace{-2mm}
Voxel-Only is essentially a version of our method that omits the mesh refinement branch,
while Pixel2Mesh$^+$ and Sphere-Init omit the voxel prediction branch.

\begin{figure}
  \centering
  \begin{minipage}{0.135\textwidth}\centering Input Image \end{minipage}
  \begin{minipage}{0.135\textwidth}\centering Without $\mathcal{L}_{edge}$ (best) \end{minipage}
  \begin{minipage}{0.135\textwidth}\centering With $\mathcal{L}_{edge}$ (pretty) \end{minipage} \\*
  % {left lower right upper}
  \includegraphics[trim={20px 0px 5px 30px},clip,width=0.135\textwidth]{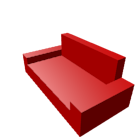}
  \includegraphics[trim={75px 0px 19px 112px},clip,width=0.135\textwidth]{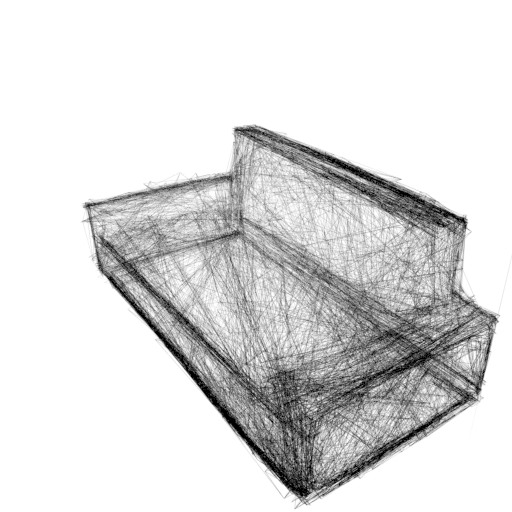}
  \includegraphics[trim={75px 0px 19px 112px},clip,width=0.135\textwidth]{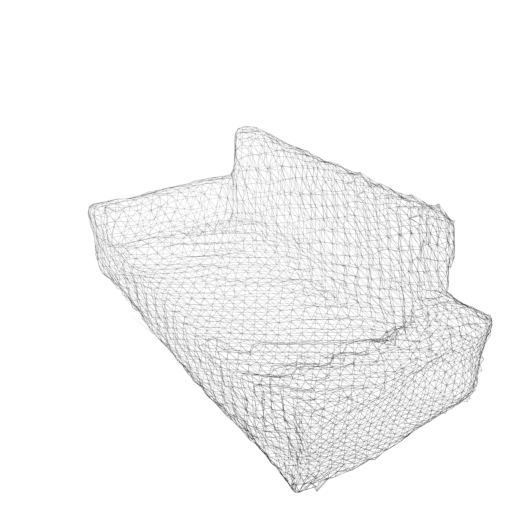} \\
  \includegraphics[trim={20px, 25px, 20px, 30px},clip,width=0.135\textwidth]{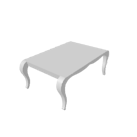}
  \includegraphics[trim={75px, 93px, 75px, 112px},clip,width=0.135\textwidth]{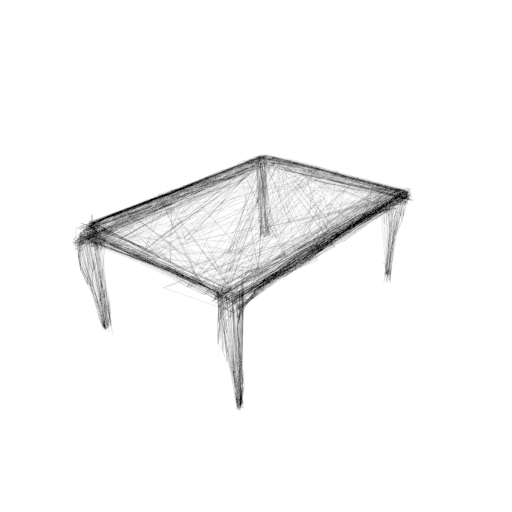}
  \includegraphics[trim={75px, 93px, 75px, 112px},clip,width=0.135\textwidth]{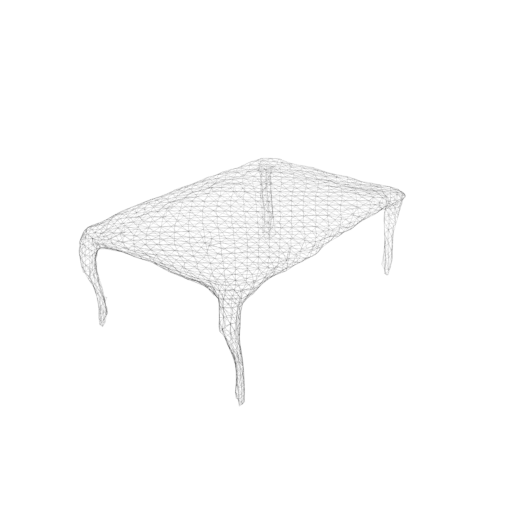}
  \caption{
    Training without the edge length regularizer $\mathcal{L}_{edge}$ results in degenerate
    predicted meshes that have many overlapping faces. Adding $\mathcal{L}_{edge}$ eliminates
    this degeneracy but results in worse agreement with the ground-truth as measured by standard
    metrics such as Chamfer distance.
  }
  \label{fig:best-vs-pretty}
  \vspace{-4mm}
\end{figure}

\ourpar{Best vs Pretty}
As previously noted in \cite{wang2018pixel2mesh} (Section 4.1), standard metrics for
shape reconstruction are not well-correlated with mesh quality.
Figure~\ref{fig:best-vs-pretty} shows that models trained without shape regularizers give
meshes that are preferred by metrics despite being highly degenerate, with irregularly-sized
faces and many self-intersections. These degenerate meshes would be difficult to texture,
and may not be useful for downstream applications.

Due to the strong effect of shape regularizers on both mesh quality and quantitative metrics,
we suggest only quantitatively comparing methods trained with the same shape regularizers.
We thus train two versions of all our ShapeNet models: a
\emph{Best} version with $\lambda_{\textrm{edge}}=0$ to serve as an upper bound on quantitative
performance, and a \emph{Pretty} version that strikes a balance
between quantitative performance and mesh quality by setting
$\lambda_{\textrm{edge}}=0.2$.

\ourpar{Comparison with Prior Work}
Table~\ref{table:shapenet-main} compares our \emph{Pretty} and \emph{Best} models with
prior work on shape prediction from a single image.
 We use the evaluation protocol from~\cite{wang2018pixel2mesh}, using a 0.57 mesh scaling factor and threshold value $\tau\hspace{-1.2mm}=\hspace{-1.2mm}10^{-4}$ 
on squared Euclidean distances. For Pixel2Mesh, we provide the performance reported in their paper~\cite{wang2018pixel2mesh} 
as well as the performance of their open-source pretrained model.
Table~\ref{table:shapenet-main} shows that we outperform prior work by a wide margin, validating the design of our mesh predictor. 

\begin{table*}
  \centering
  \tablestyle{2.5pt}{1.1}
  \begin{tabular}{l|aac|ccccccccc|TT} 
     Pix3D $\mathcal{S}_1$ & AP$^{\text{box}}$  & AP$^{\text{mask}}$ & AP$^{\text{mesh}}$ & \emph{chair} & \emph{sofa} & \emph{table}   & \emph{bed}  & \emph{desk} & \emph{bkcs} &  \emph{wrdrb} &\emph{tool} & \emph{misc}  & $|V|$ & $|F|$ \\ 
     \shline                                       
      Voxel-Only 			   & 94.4 & 88.4 & 5.3        & 0.0 		  & 3.5 		  & 2.6 		  & 0.5 		  & 0.7 		  & 34.3 	      & 5.7 		  & 0.0 	     & 0.0 			& $2354_{\pm706}$ & $4717_{\pm1423}$ \\ % f162829546
      Pixel2Mesh$^+$     & 93.5 & 88.4 & 39.9 			& 30.9 		  & 59.1 		  & 40.2 		  & 40.5 		  & 30.2 		  & 50.8 	      & 62.4 		  & 18.2 	     & 26.7 		& $2562_{\pm 0}$ & $5120_{\pm0}$  \\ % f162827706
      Sphere-Init        & 94.1 & 87.5 & 40.5 			& 40.9 		  & \bf{75.2} & 44.2 		  & 50.3      & 28.4 		  & 48.6 	      & 42.5 		  & \bf{26.9}  & 7.0 		  & $2562_{\pm 0}$ & $5120_{\pm0}$ \\ % f162828471
      Mesh R-CNN (ours)  & 94.0 & 88.4 & \bf{51.1} 	& \bf{48.2} & 71.7      & \bf{60.9} &\bf{53.7}  & \bf{42.9} & \bf{70.2}   & \bf{63.4} & 21.6       & \bf{27.8}& $2367_{\pm698}$ & $4743_{\pm1406}$ \\ % f163298658
      \hline
      {\color{Gray2} \# test instances}  &  {\color{Gray2} 2530} &  {\color{Gray2} 2530} &  {\color{Gray2} 2530} &  {\color{Gray2} 1165} &  {\color{Gray2} 415} &  {\color{Gray2} 419} &  {\color{Gray2} 213} &  {\color{Gray2} 154} &  {\color{Gray2} 79}  &  {\color{Gray2} 54} &  {\color{Gray2} 11} &  {\color{Gray2} 20}  &  & \\ 
      \vspace{-2.5mm}
      \\
      Pix3D $\mathcal{S}_2$ \\ 
      \shline
      Voxel-Only         & 71.5 & 63.4 & 4.9        & 0.0       & 0.1       &  2.5      & 2.4       & 0.8       & 32.2      & 0.0      & 6.0       & 0.0  & $2346_{\pm630}$ & $4702_{\pm1269}$    \\ % f162830497
      Pixel2Mesh$^+$     & 71.1 & 63.4 & 21.1       & 26.7 		  & 58.5		  & 10.9 		  & 38.5 	    & 7.8 		  & 34.1	    & \bf{3.4} & \bf{10.0} 		 & 0.0 	& $2562_{\pm 0}$ & $5120_{\pm0}$ \\ % f162827832
      Sphere-Init        & 72.6 & 64.5 & 24.6 		  & 32.9 		  & \bf{75.3} & 15.8 		  & 40.1      & 10.1 		  & 45.0 		  & 1.5 		 & 0.8       & 0.0 	& $2562_{\pm 0}$ & $5120_{\pm0}$ \\ % f162828827
       Mesh R-CNN (ours) & 72.2 & 63.9 & \bf{28.8} 	& \bf{42.7} & 70.8      & \bf{27.2} & \bf{40.9} & \bf{18.2} & \bf{51.1} & 2.9      & 5.2 	     & 0.0	& $2358_{\pm633}$ & $4726_{\pm1274}$  \\ % f162959878
      \hline 
      {\color{Gray2} \# test instances}  &  {\color{Gray2} 2356} &  {\color{Gray2} 2356} &  {\color{Gray2} 2356} &  {\color{Gray2} 777} &  {\color{Gray2} 504} &  {\color{Gray2} 392} &  {\color{Gray2} 218} &  {\color{Gray2} 205} &  {\color{Gray2} 84}  &  {\color{Gray2} 134} &  {\color{Gray2} 22} &  {\color{Gray2} 20}  & & \\ 
    \end{tabular}
  \vspace{2mm}
  \caption{Performance on Pix3D $\mathcal{S}_1$ \& $\mathcal{S}_2$. We report mean AP$^{\text{box}}$, AP$^{\text{mask}}$ and AP$^{\text{mesh}}$, 
  	      as well as per category AP$^{\text{mesh}}$. All AP performances are in \%.  The \emph{Voxel-Only} baseline outputs the cubified voxel predictions. The \emph{Sphere-Init} and \emph{Pixel2Mesh$^+$} baselines deform an initial sphere and thus are limited to making predictions homeomorphic to spheres. Our Mesh R-CNN is flexible and can capture arbitrary topologies. We outperform the baselines consistently while predicting meshes with fewer number of vertices and faces.
  }
   \label{table:pix3d-main}
   \vspace{-4mm}
\end{table*}

\begin{table}
  \centering
  \tablestyle{2.0pt}{1.1}
  \begin{tabular}{cc|ccc} 
       \hspace{0.1cm} CNN init  \hspace{0.1cm} &  \hspace{0.1cm} \# refine steps \hspace{0.1cm}  & AP$^{\text{box}}$  & AP$^{\text{mask}}$ & AP$^{\text{mesh}}$  \\ 
     \shline
        COCO & 3 & 94.0 & 88.4 & 51.1 \\ % f163298658
        IN 	 & 3 & 93.1 & 87.0 & 48.4 \\ % f162962970
        COCO & 2 & 94.6 & 88.3 & 49.3 \\ % f162960568
        COCO & 1 & 94.2 & 88.9 & 48.6 \\ % f162960389
     \end{tabular} 
  \vspace{2mm}
  \caption{Ablations of Mesh R-CNN on Pix3D.}
  \label{table:pix3d-ablation}
  \vspace{-4mm}
\end{table}

\begin{figure}
  \centering
  \rotatebox{90}{\hspace{6mm}Image} \hspace{2mm}
  % {left lower right upper}
  \includegraphics[trim={ 25px  15px  35px  32px},clip,width=0.08\textwidth]{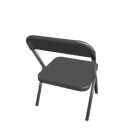} \hspace{2mm}
  \includegraphics[trim={ 28px  18px  30px  25px},clip,width=0.08\textwidth]{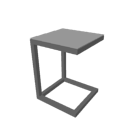} \hspace{2mm}
  \includegraphics[trim={ 20px  15px  18px   8px},clip,width=0.08\textwidth]{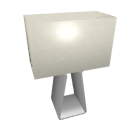}  \hspace{2mm}
  \includegraphics[trim={ 25px  25px  22px  30px},clip,width=0.08\textwidth]{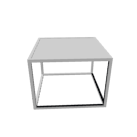} \\
  \rotatebox{90}{\hspace{1mm}\small Pixel2Mesh$^+$ \hspace{-2mm}} \hspace{2mm}
  \includegraphics[trim={ 93px  56px 131px 120px},clip,width=0.08\textwidth]{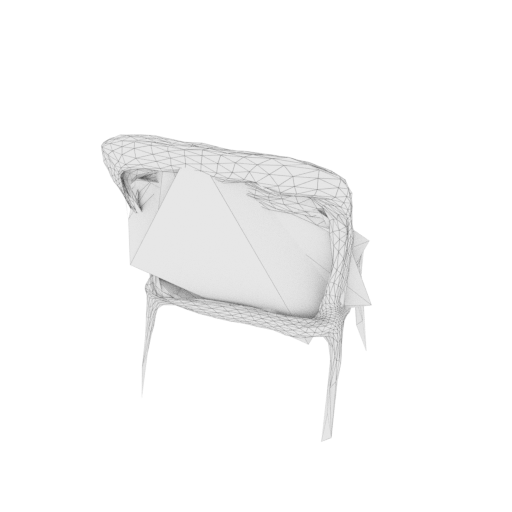} \hspace{2mm}
  \includegraphics[trim={105px  67px 112px  93px},clip,width=0.08\textwidth]{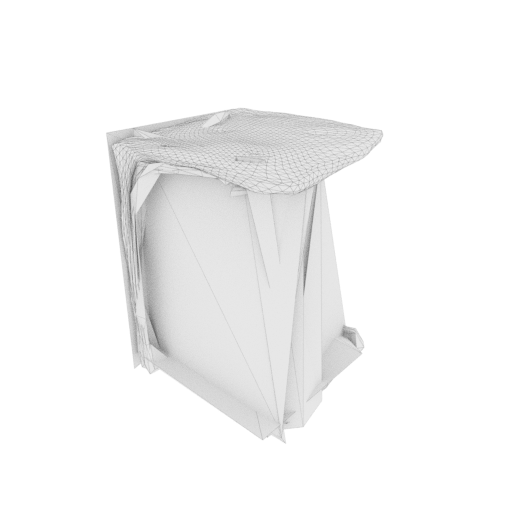} \hspace{2mm}
  \includegraphics[trim={ 75px  56px  67px  30px},clip,width=0.08\textwidth]{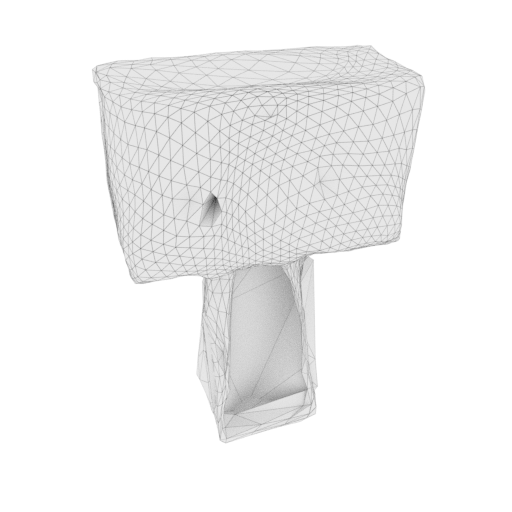}  \hspace{2mm}
  \includegraphics[trim={ 93px  93px  82px 112px},clip,width=0.08\textwidth]{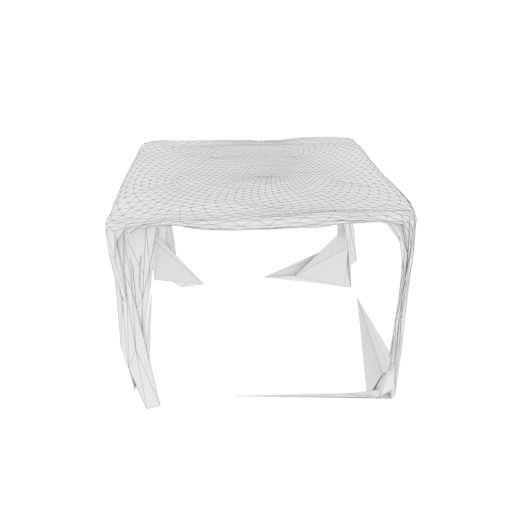} \\
  \rotatebox{90}{\hspace{7mm}Ours} \hspace{2mm}
  \includegraphics[trim={ 93px  56px 131px 120px},clip,width=0.08\textwidth]{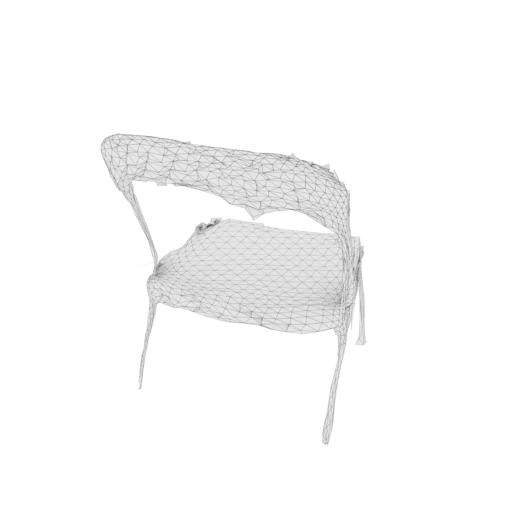} \hspace{2mm}
  \includegraphics[trim={105px  67px 112px  93px},clip,width=0.08\textwidth]{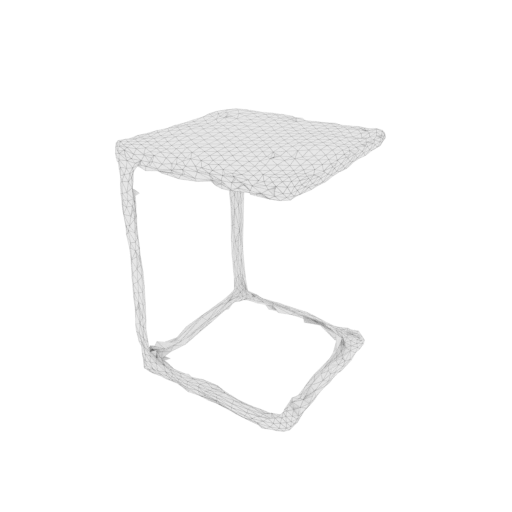} \hspace{2mm}
  \includegraphics[trim={ 75px  56px  67px  30px},clip,width=0.08\textwidth]{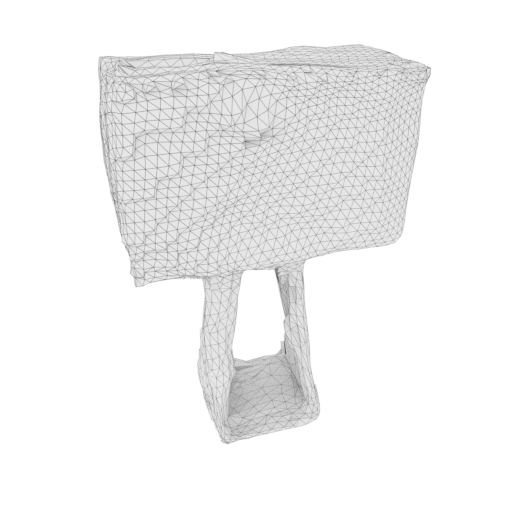}  \hspace{2mm}
  \includegraphics[trim={ 93px  93px  82px 112px},clip,width=0.08\textwidth]{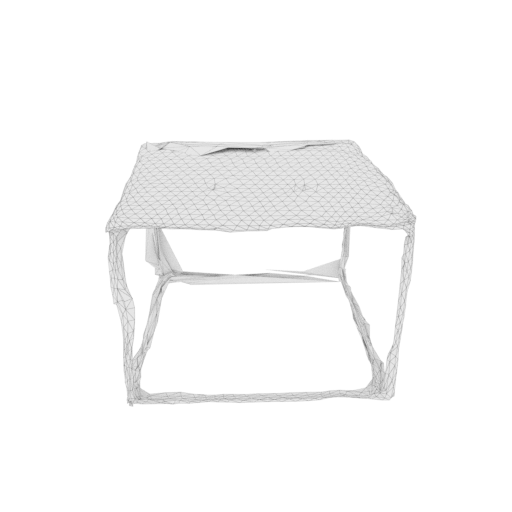}
  \caption{
    Pixel2Mesh$^+$ predicts meshes by deforming an initial sphere, so it cannot properly
    model objects with holes. In contrast our method can model objects with arbitrary topologies.
  }
  \label{fig:shapenet-holes}
  \vspace{-4mm}
\end{figure}

\ourpar{Ablation Study}
Fairly comparing with prior work is challenging due to differences in 
backbone networks, losses, and shape regularizers. For a controlled evaluation, 
we ablate variants using the same backbone and training recipe, shown in Table~\ref{table:shapenet-ablation}. 
ShapeNet is dominated by simple objects of genus zero.
Therefore we evaluate both on the entire test set and on a subset consisting of objects with one or more holes (\textbf{Holes Test Set}) \footnote{We annotated 3075 test set models and flagged whether they contained holes. This resulted in 17\% (or 534) of the models being flagged. See Appendix~\ref{app:holes} for more details and examples.}.
In this evaluation we remove the ad-hoc scaling factor of 0.57, and we rescale meshes so the longest edge of the ground-truth mesh's bounding box has length 10, following \cite{fan2017point}.
We compare the open-source Pixel2Mesh model against our ablations in this evaluation setting.
Pixel2Mesh$^+$ (our reimplementation of~\cite{wang2018pixel2mesh})
significantly outperforms the original due to an improved training recipe and deeper backbone.

We draw several conclusions from Table~\ref{table:shapenet-ablation}:
(a) On the Full Test Set, our full model and Pixel2Mesh$^+$ perform on par. However, on the Holes Test Set, our model dominates as it is able to predict topologically diverse shapes while Pixel2Mesh$^+$ is restricted to make predictions homeomorphic to spheres, and cannot model holes or disconnected components (see Figure~\ref{fig:shapenet-holes}).
This discrepancy is quantitatively more salient on Pix3D (\secref{pix3d}) as it contains more complex shapes. % with holes and disconnected components.
(b) Sphere-Init and Pixel2Mesh$^+$ perform similarly overall (both \emph{Best} and \emph{Pretty}), suggesting that mesh subdivision may be unnecessary for strong quantitative performance. % They also outperform Pixel2Mesh, showing the effectiveness of our training recipe.
(c) The deeper residual mesh refinement architecture (inspired by \cite{wang2018pixel2mesh}) performs on-par with the lighter non-residual architecture, motivating our use of the latter on Pix3D.
(d) Voxel-Only performs poorly compared to methods that predict meshes, demonstrating that mesh predictions better capture fine object structure.
(e) Each \emph{Best} model outperforms its corresponding \emph{Pretty} model; this is expected since \emph{Best} is an upper bound on quantitative performance.

\begin{figure*}
  \centering
  \includegraphics[width=\linewidth]{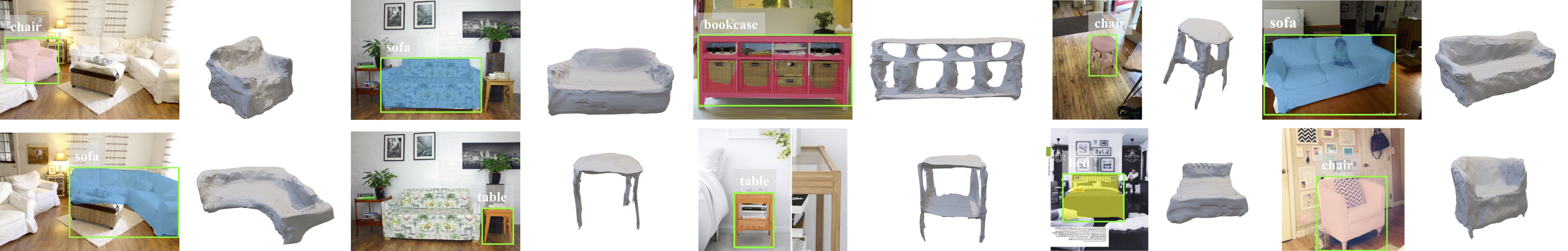}
  \caption{Examples of Mesh R-CNN predictions on Pix3D. Mesh R-CNN detects multiple objects per image, reconstructs fine details such as chair legs, and predicts varying and
complex mesh topologies for objects with holes such as bookcases and tables.}
   \label{fig:pix3d-comp}
   \vspace{-2mm}
\end{figure*}

\subsection{Pix3D}
\seclabel{pix3d}
 
We now turn to Pix3D~\cite{pix3d}, which consists of 10069 real-world images and 395 unique 3D models.
Here the task is to jointly detect and predict 3D shapes for known object categories.
Pix3D does not provide standard train/test splits, so we prepare two splits of our own.

Our first split, $\mathcal{S}_1$, randomly allocates 7539 images for training and 2530 for testing.
Despite the small number of unique object models compared to ShapeNet, $\mathcal{S}_1$ is challenging
since the same model can appear with varying appearance (\eg color, texture), in different orientations,
under different lighting conditions, in different contexts, and with varying occlusion. This is a
stark contrast with ShapeNet, where objects appear against blank backgrounds.

Our second split, $\mathcal{S}_2$, is even more challenging: we ensure that the 3D models appearing
in the train and test sets are disjoint. Success on this split requires generalization not
only to the variations present in $\mathcal{S}_1$, but also to novel 3D shapes of known categories:
for example a model may see kitchen chairs during training but must recognize armchairs during testing.
This split is possible due to Pix3D's unique annotation structure, and poses 
interesting challenges for both 2D recognition and 3D shape prediction.

\ourpar{Evaluation}
We adopt metrics inspired by those used for 2D recognition:
AP$^{\text{box}}$, AP$^{\text{mask}}$ and AP$^{\text{mesh}}$.
The first two are standard metrics used for evaluating COCO object detection and instance segmentation
at intersection-over-union (IoU) 0.5.
AP$^{\text{mesh}}$ evalutes 3D shape prediction: it is the mean area under the per-category precision-recall curves
for \Fone{0.3} at 0.5\footnote{A mesh prediction is considered a true-positive if its predicted label is correct, it is not a duplicate detection, and its F1$^{0.3}>0.5$}.
Pix3D is not exhaustively annotated, so for evaluation we only consider predictions with box $\textrm{IoU}>0.3$
with a ground-truth region. This avoids penalizing the model for correct predictions corresponding to unannotated
objects.

We compare predicted and ground-truth meshes in the \emph{camera coordinate system}.
Our model assumes known camera intrinsics for \texttt{VertAlign}.
In addition to predicting the box of each object on the image plane, Mesh R-CNN predicts the depth extent by appending a 2-layer MLP head, similar to the box regressor head. As a result, Mesh R-CNN predicts a 3D bounding box for each object. See Appendix~\ref{app:zpred} for more details.

\ourpar{Implementation details} 
We use ResNet-50-FPN~\cite{Lin2017fpn} as the backbone CNN; the box and mask branches are identical to Mask R-CNN.
The voxel branch resembles the mask branch, but the pooling resolution is decreased to 12
(\vs 14 for masks) due to memory constraints giving $24\times24\times24$ voxel predictions.
We adopt the lightweight design for the mesh refinement branch from~\secref{shapenet}.
We train for 12 epochs with a batch size of 64 per image on 8 Tesla V100 GPUs (two images per GPU).
We use SGD with momentum, linearly increasing the learning rate from $0.002$ to $0.02$ over the first 1K
iterations, then decaying by a factor of 10 at 8K and 10K iterations.
We initialize from a model pretrained for instance segmentation on COCO.
We set the \texttt{cubify} threshold to $0.2$ and the loss weights to
$\lambda_{\textrm{voxel}}=3$, $\lambda_{\textrm{cham}}=1$, $\lambda_{\textrm{norm}} = 0.1$ and
$\lambda_{\textrm{edge}}=1$ and use weight decay $10^{-4}$;
detection loss weights are identical to Mask R-CNN.

\ourpar{Comparison to Baselines}
As discussed in \secref{sec:Intro}, we are the first to tackle joint detection and shape inference
\emph{in the wild} on Pix3D. To validate our approach we compare with ablated versions of Mesh R-CNN, replacing
our full mesh predictor with \emph{Voxel-Only}, \emph{Pixel2Mesh}$^+$, and \emph{Sphere-Init} branches
(see \secref{shapenet}). All baselines otherwise use the same architecture and training recipe.

Table~\ref{table:pix3d-main} (top) shows the performance on $\mathcal{S}_1$. We observe that:
(a) Mesh R-CNN outperforms all baselines, improving over the next-best by 10.6\% \APmesh overall and across most categories;
\emph{Tool} and \emph{Misc}\footnote{\emph{Misc} consists of objects such as fire hydrant, picture frame, vase, \etc}
have very few test-set instances (11 and 20 respectively), so their AP is noisy.
(b) Mesh R-CNN shows large gains \vs \emph{Sphere-Init} for objects with complex shapes such as \emph{bookcase} (+21.6\%), \emph{table} (+16.7\%) and \emph{chair} (+7.3\%).
(c) Voxel-Only performs very poorly -- this is expected due to its coarse predictions.

Table~\ref{table:pix3d-main} (bottom) shows the performance on  the more challenging $\mathcal{S}_2$ split. Here we observe: (a) The overall performance on 2D recognition (\APbox, \APmask) drops significantly compared to $\mathcal{S}_1$, signifying the difficulty of recognizing novel shapes in the wild. (b) Mesh R-CNN outperforms all baselines for shape prediction for all categories except \emph{sofa}, \emph{wardrobe} and \emph{tool}. (c) Absolute performance on \emph{wardrobe}, \emph{tool} and \emph{misc} is small for all methods due to significant shape disparity between models in train and test and lack of training data.

Table~\ref{table:pix3d-ablation} compares pretraining on COCO vs ImageNet, and compares different architectures for the mesh
predictor. COCO \vs ImageNet initialization improves 2D recognition (\APmask 88.4 \vs 87.0) and  3D shape prediction (\APmesh 51.1 \vs 48.4). 
Shape prediction is degraded when using only one mesh refinement stage (\APmesh 51.1 \vs 48.6).

\begin{figure}
  \centering
  \includegraphics[width=\linewidth]{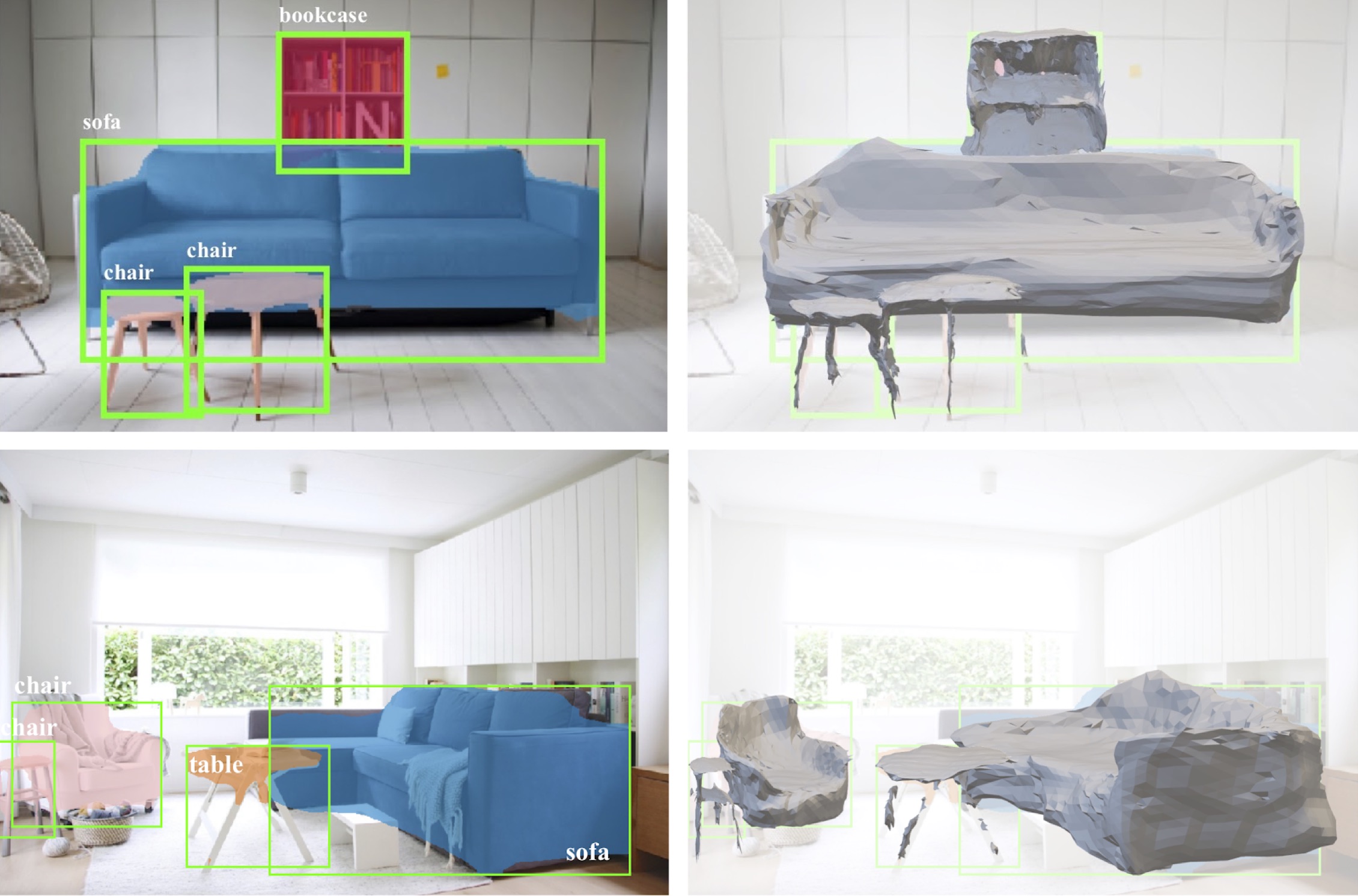}
  \caption{More examples of Mesh R-CNN predictions on Pix3D.}
   \label{fig:pix3d-scenep}
   \vspace{-2mm}
\end{figure}

Figures~\ref{fig:preds-teaser}, \ref{fig:pix3d-comp} and \ref{fig:pix3d-scenep} show example predictions from Mesh R-CNN.
Our method can detect multiple objects per image, reconstruct fine details such as chair legs, and predict varying and
complex mesh topologies for objects with holes such as bookcases and desks.

\section*{Discussion}

We propose Mesh R-CNN, a novel system for joint 2D perception and 3D shape inference. We validate our approach on ShapeNet and show its merits on Pix3D. Mesh R-CNN is a first attempt at 3D shape prediction in the wild. Despite the lack of large supervised data, \eg compared to COCO, Mesh R-CNN shows promising results. Mesh R-CNN is an object centric approach. Future work includes reasoning about the 3D layout, \ie the relative pose of objects in the 3D scene. 

\ourpar{Acknowledgements} 
We would like to thank Kaiming He, Piotr Doll\'ar, Leonidas Guibas, Manolis Savva and Shubham Tulsiani for valuable discussions. We would also like to thank Lars Mescheder and Thibault Groueix for their help.
\section*{Appendix}
\appendix

% -------------- Cubify ------------ %
\section{Implementation of \texttt{Cubify}}
\label{app:cubify}
Algorithm~\ref{arg:cubify} outlines the \texttt{cubify} operation. \texttt{Cubify} takes as input voxel occupancy probabilities $V$ of shape $N \times D \times H \times W$ as predicted from the voxel branch and a threshold value $\tau$. Each occupied voxel is replaced with a cuboid triangle mesh ($\textrm{unit\_cube}$) with
8 vertices, 18 edges, and 12 faces. Shared vertices and edges between adjacent occupied voxels are merged, and shared interior faces are eliminated. 
This results in a watertight mesh $T=(V,F)$ for each example in the batch whose topology depends on the voxel predictions.

Algorithm~\ref{arg:cubify} is an inefficient implementation of \texttt{cubify} as it involves nested for loops which in practice increase the time complexity, especially for large batches and large voxel sizes. In particular, this implementation takes $>$~300ms for $N=32$ voxels of size $32\times32\times32$ on a Tesla V100 GPU. We replace the nested for loops with 3D convolutions and vectorize our computations, resulting in a time complexity of $\approx$ 30ms for the same voxel inputs.

\begin{algorithm}
\SetAlgoLined
\KwData{$V:[0,1]^{N \times D \times H \times W}$, $\tau \in [0,1]$}
 $\textrm{unit\_cube} = (V_{\textrm{cube}}, F_{\textrm{cube}})$\\
 \For{(n, z, y, x) $\in$ range(N, D, H, W)}{
 	\If{$V[n, z, y, x] > \tau$}{
    		add $\textrm{unit\_cube}$ at $(n,z,y,x)$\\
		\If{$V[n, z-1, y, x] > \tau$}{
			remove back faces\\
		} 
		\If{$V[n, z+1, y, x] > \tau$}{
			remove front faces\\
		} 
		\If{$V[n, z, y-1, x] > \tau$}{
			remove top faces\\
		} 
		\If{$V[n, z, y+1, x] > \tau$}{
			remove bottom faces\\
		} 
		\If{$V[n, z, y, x-1] > \tau$}{
			remove left faces\\
		} 
		\If{$V[n, z, y, x+1] > \tau$}{
			remove right faces\\
		} 
	}
 }
 merge shared verts\\
 \Return A list of meshes $\{T_i = (V_i, F_i)\}_{i=0}^{N-1}$
 \vspace{2mm}
 \caption{\texttt{Cubify}}
 \label{arg:cubify}
\end{algorithm}

% -------------- Mesh Sampling ------------ %
\section{Mesh Sampling}
\label{app:meshsampling}
As described in the main paper, the mesh refinement head is trained to minimize \emph{chamfer} and \emph{normal} losses that are defined on sets of points sampled from the predicted and ground-truth meshes.

Computing these losses requires some method of converting meshes into sets of sampled points.
Pixel2Mesh \cite{wang2018pixel2mesh} is trained using similar losses. In their case ground-truth meshes are represented with points sampled uniformly at random from the surface of the mesh, but this sampling is performed offline before the start of training; they represent predicted meshes using their vertex positions. Computing these losses using vertex positions of predicted meshes is very efficient since it avoids the need to sample meshes online during training; however it can lead to degenerate predictions since the loss would not encourage the interior of predicted faces to align with the ground-truth mesh.

To avoid these potential degeneracies, we follow \cite{smith2019geometrics} and compute the chamfer and normal losses by randomly sampling points from both the predicted and ground-truth meshes. This means that we need to sample the predicted meshes online during training, so the sampling must be efficient and we must be able to backpropagate through the sampling procedure to propagate gradients backward from the sampled points to the predicted vertex positions.

Given a mesh with vertices $V\subset\mathbb{R}^3$ and faces $F\subseteq V\times V\times V$, we can sample a point uniformly from the surface of the mesh as follows.
We first define a probability distribution over faces where each face's probability is proportional to its area:
\begin{equation}
  P(f) = \frac{area(f)}{\sum_{f'\in F}area(f')}
\end{equation}
We then sample a face $f=(v_1, v_2, v_3)$ from this distribution. Next we sample a point $p$ uniformly from the interior of $f$ by setting $p=\sum_iw_iv_i$ where $w_1=1-\sqrt{\xi_1}$, $w_2=(1-\xi_2)\sqrt\xi_1$, $w_3=\xi_2\sqrt\xi_1$, and $\xi_1,\xi_2\sim U(0,1)$ are sampled from a uniform distribution.

This formulation allows propagating gradients from $p$ backward to the face vertices $v_i$ and can be seen as an instance of the reparameterization trick~\cite{kingma2014auto}.

% -------------- Pix3D Architecture ------------ %
\section{Mesh R-CNN Architecture}
\label{app:arch}
At a high level we use the same overall architecture for predicting meshes on ShapeNet and Pix3D,
but we slightly specialize to each dataset due to memory constraints from the backbone and task-specific heads. On Pix3D Mask R-CNN adds time and memory complexity in order to perform object detection and instance segmentation.

\begin{table}
  \centering
  \scalebox{0.75}{
  \begin{tabular}{|c|c|l|c|}
    \hline
    \textbf{Index} & \textbf{Inputs} & \textbf{Operation} & \textbf{Output shape} \\
    \hline
    (1) & Input & \texttt{conv2\_3} features & $35\times35\times256$ \\
    (2) & Input & \texttt{conv3\_4} features & $18\times18\times512$ \\
    (3) & Input & \texttt{conv4\_6} features & $9\times9\times1024$ \\
    (4) & Input & \texttt{conv5\_3} features & $5\times5\times2048$ \\
    (5) & Input & Input vertex features & $|V|\times128$ \\
    (6) & Input & Input vertex positions & $|V|\times3$ \\
    (7) & (1), (6) & \texttt{VertAlign} & $|V|\times256$ \\
    (8) & (2), (6) & \texttt{VertAlign} & $|V|\times512$ \\
    (9) & (3), (6) & \texttt{VertAlign} & $|V|\times1024$ \\
    (10) & (4), (6) & \texttt{VertAlign} & $|V|\times2048$ \\
    (11) & (7),(8),(9),(10) & Concatenate & $|V|\times3840$ \\
    (12) & (11) & Linear$(3840\rightarrow128)$ & $|V|\times128$ \\
    (13) & (5), (6), (12) & Concatenate & $|V|\times259$ \\
    (14) & (13) & ResGraphConv$(259\rightarrow128)$ & $|V|\times128$ \\
    (15) & (14) & $2\times$ ResGraphConv$(128\rightarrow128)$ & $|V|\times128$ \\
    (16) & (15) & GraphConv$(128\rightarrow3)$ & $|V|\times3$ \\
    (17) & (16) & Tanh & $|V|\times3$ \\
    (18) & (6), (17) & Addition & $|V|\times3$ \\
    \hline
  \end{tabular}}
  \vspace{2mm}
  \caption{
    Architecture for a single residual mesh refinement stage on ShapeNet.
    For ShapeNet we follow \cite{wang2018pixel2mesh} and use residual blocks of
    graph convolutions: ResGraphConv$(D_1\rightarrow D_2)$ consists of two
    graph convolution layers (each preceeded by ReLU) and an additive skip
    connection, with a linear projection if the input and output dimensions
    are different. The output of the refinement stage are the vertex features
    (15) and the updated vertex positions (18). The first refinement stage
    does not take input vertex features (5), so for this stage (13) only
    concatenates (6) and (12).
  }
  \label{table:shapenet-refinement}
  \vspace{-2mm}
\end{table}

\begin{table}
  \vspace{1mm}
  \centering
  \scalebox{0.75}{
  \begin{tabular}{|c|c|l|c|}
    \hline
    \textbf{Index} & \textbf{Inputs} & \textbf{Operation} & \textbf{Output shape} \\
    \hline
    (1) & Input & Image & $137 \times 137 \times 3$ \\
    (2) & (1) & ResNet-50 \texttt{conv2\_3} & $35\times35\times256$ \\
    (3) & (2) & ResNet-50 \texttt{conv3\_4} & $18\times18\times512$ \\
    (4) & (3) & ResNet-50 \texttt{conv4\_6} & $9\times9\times1024$ \\
    (5) & (4) & ResNet-50 \texttt{conv5\_3} & $5\times5\times2048$ \\
    (6) & (5) & Bilinear interpolation & $24\times24\times2048$ \\
    (7) & (6) & Voxel Branch & $48\times48\times48$ \\
    (8) & (7) & \texttt{cubify} & $|V|\times3, |F|\times 3$ \\
    (9) & (2), (3), (4), (5),  (8) & Refinement Stage 1 & $|V|\times3, |F|\times 3$ \\
    (10) & (2), (3), (4), (5),  (9) & Refinement Stage 2 & $|V|\times3, |F|\times 3$ \\
    (11) & (2), (3), (4), (5), (10) & Refinement Stage 3 & $|V|\times3, |F|\times 3$ \\
    \hline
  \end{tabular}}
  \vspace{2mm}
  \caption{
    Overall architecture for our ShapeNet model. Since we do not predict bounding boxes
    or masks, we feed the \texttt{conv5\_3} features from the whole image into the
    voxel branch. The architecture for the refinement stage is shown in
    Table~\ref{table:shapenet-refinement}, and the architecture for the voxel
    branch is shown in Table~\ref{table:vox-branch}. 
  }
  \label{table:shapenet-arch}
  \vspace{-2mm}
\end{table}

\vspace{1mm}\noindent\textbf{ShapeNet.}
The overall architecture of our ShapeNet model is shown in Table~\ref{table:shapenet-arch};
the architecture of the voxel branch is shown in Table~\ref{table:vox-branch}.

We consider two different architectures for the mesh refinement network on ShapeNet.
Our full model as well as our Pixel2Mesh$^+$ and Sphere-Init baselines use mesh refinement stages
with three residual blocks of two graph convolutions each, similar to \cite{wang2018pixel2mesh};
the architecture of these stages is shown in Table~\ref{table:shapenet-refinement}.
We also consider a shallower lightweight design which uses only three graph convolution layers
per stage, omitting residual connections and instead concatenating the input vertex positions before each
graph convolution layer. The architecture of this lightweight design is shown in Table~\ref{table:shapenet-light}.

As shown in Table~\ref{table:shapenet-ablation}, we found that these two architectures perform similarly on
ShapeNet even though the lightweight design uses half as many graph convolution layers per stage.
We therefore use the nonresidual design for our Pix3D models.

\vspace{1mm}\noindent\textbf{Pix3D.}
The overall architecture of our full Mesh R-CNN system on Pix3D is shown in Table~\ref{table:pix3d-arch}.
The backbone, RPN, box branch, and mask branch are identical to Mask R-CNN~\cite{he2017maskrcnn}.
The voxel branch is the same as in the ShapeNet models (see Table~\ref{table:vox-branch}),
except that we predict voxels at a lower resolution
($48\times48\times48$ for ShapeNet \vs $24\times24\times24$ for Pix3D) due to memory constraints.
Table~\ref{table:pix3d-refinement} shows the exact architecture of the mesh refinement stages for our Pix3D models.

\begin{table}
  \centering
  \scalebox{0.75}{
  \begin{tabular}{|c|c|l|c|}
    \hline
    \textbf{Index} & \textbf{Inputs} & \textbf{Operation} & \textbf{Output shape} \\
    \hline
    (1) & Input & \texttt{conv2\_3} features & $35\times35\times256$ \\
    (2) & Input & \texttt{conv3\_4} features & $18\times18\times512$ \\
    (3) & Input & \texttt{conv4\_6} features & $9\times9\times1024$ \\
    (4) & Input & \texttt{conv5\_3} features & $5\times5\times2048$ \\
    (5) & Input & Input vertex features & $|V|\times128$ \\
    (6) & Input & Input vertex positions & $|V|\times3$ \\
    (7) & (1), (6) & \texttt{VertAlign} & $|V|\times256$ \\
    (8) & (2), (6) & \texttt{VertAlign} & $|V|\times512$ \\
    (9) & (3), (6) & \texttt{VertAlign} & $|V|\times1024$ \\
    (10) & (4), (6) & \texttt{VertAlign} & $|V|\times2048$ \\
    (11) & (7),(8),(9),(10) & Concatenate & $|V|\times3840$ \\
    (12) & (11) & Linear$(3840\rightarrow128)$ & $|V|\times128$ \\
    (13) & (5), (6), (12) & Concatenate & $|V|\times259$ \\
    (14) & (13) & GraphConv$(259\rightarrow128)$ & $|V|\times128$ \\
    (15) & (6), (14) & Concatenate & $|V|\times131$ \\
    (16) & (15) & GraphConv$(131\rightarrow128)$ & $|V|\times128$ \\
    (17) & (6), (16) & Concatenate & $|V|\times131$ \\
    (18) & (17) & GraphConv$(131\rightarrow128)$ & $|V|\times128$ \\
    (19) & (18) & Linear$(128\rightarrow3)$ & $|V|\times3$ \\
    (20) & (19) & Tanh & $|V|\times3$ \\
    (21) & (6), (20) & Addition & $|V|\times3$ \\
    \hline
  \end{tabular}}
  \vspace{2mm}
  \caption{
    Architecture for the nonresidual mesh refinement stage use in the
    lightweight version of our ShapeNet models. Each GraphConv operation is followed
    by ReLU. The output of the stage are the vertex features (18)
    and updated vertex positions (21). The first refinement stage does not take input
    vertex features (5), so for this stage (13) only concatenates (6) and (12).
  }
  \label{table:shapenet-light}
\end{table}

\begin{table}
  \centering
  \scalebox{0.75}{
  \begin{tabular}{|c|c|l|c|}
    \hline
    \textbf{Index} & \textbf{Inputs} & \textbf{Operation} & \textbf{Output shape} \\
    \hline
    (1) & Input & Image features & $V/2\times V/2 \times D$ \\
    (2) & (1) & Conv$(D\rightarrow256, 3\times3)$, ReLU & $V/2\times V/2\times256$ \\
    (3) & (2) & Conv$(256\rightarrow256, 3\times3)$, ReLU & $V/2\times V/2\times256$ \\
    (4) & (3) & TConv$(256\rightarrow256, 2\times2, 2)$, ReLU & $V\times V\times256$ \\
    (5) & (4) & Conv$(256\rightarrow V, 1\times1)$ & $V\times V\times V$ \\
  \hline
  \end{tabular}}
  \vspace{2mm}
  \caption{
    Architecture of our voxel prediction branch. For ShapeNet we use $V=48$ and for Pix3D
    we use $V=24$. TConv is a transpose convolution with stride 2.
  }
  \label{table:vox-branch}
\end{table}

\vspace{1mm}\noindent\textbf{Baselines.}
The Voxel-Only baseline is identical to the full model, except that it omits all mesh refinement branches
and terminates with the mesh resulting from \texttt{cubify}. On ShapeNet, the Voxel-Only baseline is
trained with a batch size of 64 (\vs a batch size of 32 for our full model); on Pix3D it uses
the same training recipe as our full model.

\begin{table*}
  \centering
  \setlength{\tabcolsep}{1mm}
  \scalebox{0.88}{
  \begin{tabular}{|c|c|l|c|}
    \hline
    \textbf{Index} & \textbf{Inputs} & \textbf{Operation} & \textbf{Output shape} \\
    \hline
    (1) & Input & Input Image & $H\times W \times3$ \\
    (2) & (1) & Backbone: ResNet-50-FPN & $h\times w\times 256$ \\
    (3) & (2) & RPN & $h\times w \times A \times 4$ \\
    (4) & (2),(3) & \texttt{RoIAlign} & $14\times14\times256$ \\
    (5) & (4) & Box branch: $2\times$ downsample, Flatten, Linear$(7*7*256\rightarrow1024)$, Linear$(1024\rightarrow 5C)$ & $C\times 5$\\
    (6) & (4) & Mask branch: $4\times$ Conv$(256\rightarrow256,3\times3)$, TConv$(256\rightarrow256, 2\times2, 2)$,
                              Conv($256\rightarrow C, 1\times1$) & $28\times28\times C$ \\
    (7) & (2), (3) & \texttt{RoIAlign} & $12\times12\times256$ \\
    (8) & (7) & Voxel Branch & $24\times24\times24$ \\
    (9) & (8) & \texttt{cubify} & $|V|\times3$, $|F|\times3$ \\
    (10) & (7), (9) & Refinement Stage 1 & $|V|\times3$, $|F|\times3$ \\
    (11) & (7), (10) & Refinement Stage 2 & $|V|\times3$, $|F|\times3$ \\
    (12) & (7), (11) & Refinement Stage 3 & $|V|\times3$, $|F|\times3$ \\
    \hline
  \end{tabular}}
  \vspace{2mm}
  \caption{
    Overall architecture of Mesh R-CNN on Pix3D. The backbone, RPN, box, and mask branches are identical to Mask R-CNN.
    The RPN produces a bounding box prediction for each of the $A$ anchors at each spatial location in the input feature map;
    a subset of these candidate boxes are processed by the other branches, but here we show only the shapes resulting from
    processing a single box for the subsequent task-specific heads.
    Here $C$ is the number of categories
    (10 = 9 + \texttt{background} for Pix3D); the box branch produces per-category bounding boxes and classification scores,
    while the mask branch produces per-category segmentation masks. TConv is a transpose convolution with stride 2.
    We use a ReLU nonlinearity between all Linear, Conv, and TConv operations. The architecture fo the voxel branch
    is shown in Table~\ref{table:vox-branch}, and the architecture of the refinement stages is shown in
    Table~\ref{table:pix3d-refinement}.
  }
  \label{table:pix3d-arch}
\end{table*}

\begin{table}
  \centering
  \scalebox{0.75}{
  \begin{tabular}{|c|c|l|c|}
    \hline
    \textbf{Index} & \textbf{Inputs} & \textbf{Operation} & \textbf{Output shape} \\
    \hline
    (1) & Input & Backbone features & $h\times w\times256$ \\
    (2) & Input & Input vertex features & $|V|\times128$ \\
    (3) & Input & Input vertex positions & $|V|\times3$ \\
    (4) & (1), (3) & \texttt{VertAlign} & $|V|\times256$ \\
    (5) & (2), (3), (4) & Concatenate & $|V|\times387$ \\
    (6) & (5) & GraphConv$(387\rightarrow128)$ & $|V|\times128$ \\
    (7) & (3), (6) & Concatenate & $|V|\times131$ \\
    (8) & (7) & GraphConv$(131\rightarrow128)$ & $|V|\times128$ \\
    (9) & (3), (8) & Concatenate & $|V|\times131$ \\
    (10) & (9) & GraphConv$(131\rightarrow128)$ & $|V|\times128$ \\
    (11) & (3), (10) & Concatenate & $|V|\times 131$ \\
    (12) & (11) & Linear$(131\rightarrow3)$ & $|V|\times3$ \\
    (13) & (12) & Tanh & $|V|\times3$ \\
    (14) & (3), (13) & Addition & $|V|\times3$ \\
    \hline
  \end{tabular}}
  \vspace{2mm}
  \caption{
    Architecture for a single mesh refinement stage on Pix3D.
  }
  \label{table:pix3d-refinement}
\end{table}

The Pixel2Mesh$^+$ is our reimplementation of \cite{wang2018pixel2mesh}. This baseline omits the voxel
branch; instead all images use an identical inital mesh. The initial mesh is a level-2 icosphere with
162 vertices, 320 faces, and 480 edges which results from applying two face subdivision operations to
a regular icosahedron and projecting all resulting vertices onto a sphere. For the Pixel2Mesh$^+$
baseline, the mesh refinement stages are the same as our full model, except that we apply a face
subdivision operation prior to \texttt{VertAlign} in refinement stages 2 and 3.

Like Pixel2Mesh$^+$, the Sphere-Init baseline omits the voxel branch and uses an identical initial
sphere mesh for all images. However unlike Pixel2Mesh$^+$ the initial mesh is a level-4 icosphere
with 2562 vertices, 5120 faces, and 7680 edges which results from applying four face subdivivison
operations to a regular icosahedron. Due to this large initial mesh, the mesh refinement stages
are identical to our full model, and do not use mesh subdivision. 

Pixel2Mesh$^+$ and Sphere-Init both predict meshes with the same number of vertices and faces,
and with identical topologies; the only difference between them is whether all
subdivision operations are performed before the mesh refinement branch (Sphere-Init) or whether
mesh refinement is interleaved with mesh subdivision (Pixel2Mesh$^+$).
On ShapeNet, the Pixel2Mesh$^+$ and Sphere-Init baselines are trained with a batch size of 96;
on Pix3D they use the same training recipe as our full model.

\begin{table}
  \centering
  \tablestyle{3pt}{1.1}
  \scalebox{0.85}{
  \begin{tabular}{cc|ccccccc}
      & & Chamfer($\downarrow$) & Normal & \Fone{0.1} & \Fone{0.3} & \Fone{0.5} & $|V|$ & $|F|$ \\
    \shline
      & OccNet~\cite{mescheder2018occupancy} & 0.264 & 0.789 & 33.4 & 80.5 & 91.3
        & \meanstd{2499}{60} & \meanstd{4995}{120} \\
    \hline\hline
    \multirow{2}{*}{\rotatebox[origin=c]{90}{Best}}
      & Ours (light) & 0.135 & 0.725 & 38.9 & 86.7 & 95.0 
        & \meanstd{1978}{951} & \meanstd{3958}{1906} \\
      & Ours & 0.139 & 0.728 & 38.3 & 86.3 & 94.9
        & \meanstd{1985}{960} & \meanstd{3971}{1924} \\
    \hline\hline
    \multirow{2}{*}{\rotatebox[origin=c]{90}{Pretty}}
     & Ours (light) & 0.185 & 0.696 & 34.3 & 82.0 & 92.8
        & \meanstd{1976}{956} & \meanstd{3954}{1916} \\
     & Ours & 0.180 & 0.709 & 34.6 & 82.2 & 93.0 
        & \meanstd{1982}{961} & \meanstd{3967}{1926} \\
  \end{tabular}}
  \vspace{2mm}
  \caption{
    Comparison between our method and Occpancy Networks (OccNet)~\cite{mescheder2018occupancy}
    on ShapeNet. We use the same evaluation metrics and setup as Table~\ref{table:shapenet-ablation}.
  }
  \label{table:occnet}
\end{table}

\section{Comparison with Occupancy Networks}
\label{app:occnet}
Occupancy Networks~\cite{mescheder2018occupancy} (OccNet) also predict 2D meshes with
neural networks. Rather than outputing a mesh directly from the neural network as in our approach, they train
a neural network to compute a signed distance between a query point in 3D space and the object boundary.
At test-time a 3D mesh can be extracted from a set of query points. Like our approach, OccNets can also
predict meshes with varying topolgy per input instance.

Table~\ref{table:occnet} compares our approach with OccNet on the ShapeNet test set.
We obtained test-set predictions for OccNet from the authors.
Our method and OccNet are trained on slightly different splits of the ShapeNet dataset,
so we compare our methods on the intersection of our respective test splits.
From Table~\ref{table:occnet} we see that OccNets achieve higher normal consistency than
our approach; however both the \emph{Best} and \emph{Pretty} versions of our model outperform
OccNets on all other metrics.

% -------------- Depth Extent Prediction ------------ %
\section{Depth Extent Prediction}
\label{app:zpred}

Predicting an object's depth extent from a single image is an ill-posed problem. In an earlier version of our work, we assumed the range of an object in the $Z$-axis was given at train \& test time. Since then, we have attempted to predict the depth extent by training a 2-layer MLP head of similar architecture to the bounding box regressor head. Formally, this head is trained to predict the scale-normalized depth extent (in $\log$ space) of the object, as follows

\begin{equation}
\bar{dz} = \frac{dz}{z_c} \cdot \frac{f}{h}
\end{equation}

Note that the depth extent of an object is related to the size of the object (here approximated by the object's bounding box height $h$), its location $z_c$ in the $Z$-axis (far away objects need to be bigger in order to explain the image) and the focal length $f$. At inference time the depth extent $dz$ of the object is recovered from the predicted $\bar{dz}$ and the predicted height of the object bounding box $h$, given the focal length $f$ and the center of the object $z_c$ in the $Z$-axis. Note that we assume the center of the object $z_c$ since Pix3D annotations are not metric and due to the inherent scale-depth ambiguity. 

% -------------- Pix3D Visualizations ------------ %
\section{Pix3D: Visualizations and Comparisons}
\label{app:pix3d}

\begin{figure}
  \centering
   \includegraphics[width=\linewidth]{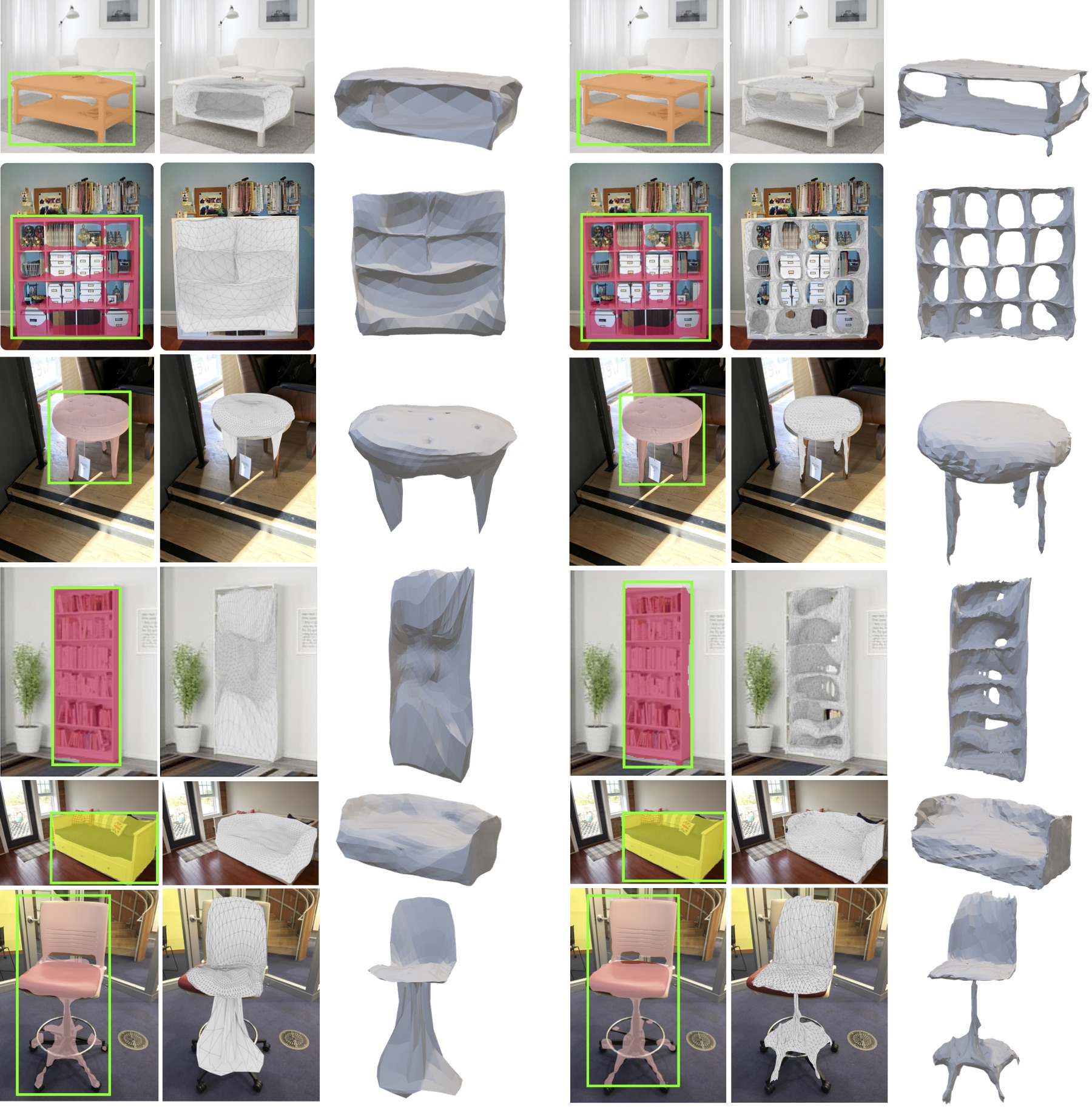}
   \begin{minipage}{\linewidth} \hspace{10mm} Pixel2Mesh$^+$ \hspace{22mm} Mesh R-CNN \end{minipage}
   \vspace{1mm}
    \caption{Qualitative comparisons between Pixel2Mesh$^+$ and Mesh R-CNN on Pix3D. Each row shows the same example for Pixel2Mesh$^+$ (first three columns) and Mesh R-CNN (last three columns), respectively. For each method, we show the input image along with the predicted 2D mask ({\color[rgb]{1,0.71,0.76} chair}, {\color[rgb]{1,0.08,0.57} bookcase}, {\color[rgb]{1,0.55,0} table}, {\color[rgb]{1,1,0} bed}) and box (in {\color{green} green}) superimposed. We show the 3D mesh rendered on the input image and an additional view of the 3D mesh.}
   \label{fig:pix3d_comp}
\end{figure}

Figure~\ref{fig:pix3d_comp} shows qualitative comparisons between Pixel2Mesh$^+$ and Mesh R-CNN. Pixel2Mesh$^+$ is limited to making predictions homeomorphic to spheres and thus cannot capture varying topologies, e.g. holes. In addition, Pixel2Mesh$^+$ has a hard time capturing high curvatures, such as sharp table tops and legs. This is due to the large deformations required when starting from a sphere, which are not encouraged by the shape regularizers. On the other hand, Mesh R-CNN initializes its shapes with cubified voxel predictions resulting in better initial shape representations which require less drastic deformations.

% -------------- ShapeNet Holes ------------ %
\section{ShapeNet Holes test set}
\label{app:holes}

\begin{figure*}
  \centering
  \input{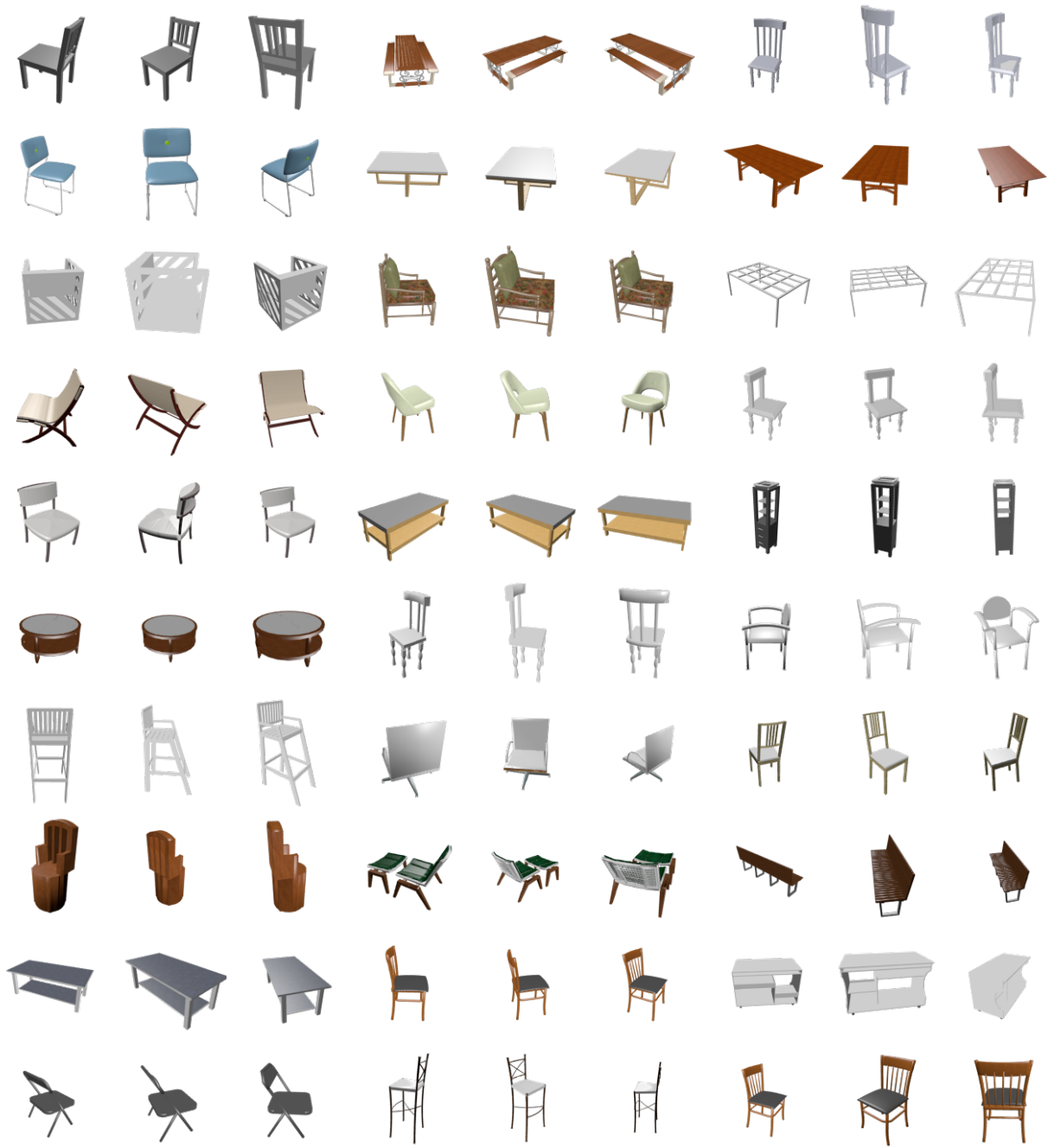}
  \caption{
    Example input images for randomly selected models from the the Holes Test Set on ShapeNet.
    For each model we show three different input images showing the model from different viewpoints.
    This set is extremely challenging -- some models may have very small holes (such as the holes in the back of the
    chair in the left model of the first row, or the holes on the underside of the table on the right model of row 2),
    and some models may have holes which are not visible in all input images (such as the green chair in the middle of
    the fourth row, or the gray desk on the right of the ninth row).
    }
    \label{fig:shapenet_holes_random}
\end{figure*}

We construct the ShapeNet Holes Test set by selecting models from the ShapeNet test set that have visible holes from any viewpoint.
Figure~\ref{fig:shapenet_holes_random} shows several input images for randomly selected models from this subset.
This test set is very challenging -- many objects have small holes resulting from thin structures; and some objects have holes which
are not visible from all viewpoints.

{\small
\bibliographystyle{ieee_fullname}
\bibliography{refs}
}

\end{document}